\documentclass[preprint,12pt]{elsarticle}

\usepackage[utf8]{inputenc}
\usepackage[a4paper, total={6in, 8in}]{geometry}
\usepackage{amsmath}
\usepackage{amssymb}
\usepackage{pgfplots}
\pgfplotsset{compat=1.15}
\usetikzlibrary{arrows}
\usepackage{algorithm} 
\usepackage{algpseudocode}
\usepackage{tabularx}
\usepackage[caption=false]{subfig}
\usepackage{lineno}
\usepackage{url}
\usepackage[splitrule]{footmisc}
    
\makeatletter
\newcommand{\multiline}[1]{%
  \begin{tabularx}{\dimexpr\linewidth-\ALG@thistlm}[t]{@{}X@{}}
    #1
  \end{tabularx}
}
\makeatother

\journal{Engineering Applications of Artificial Intelligence}

\begin{document}

\begin{frontmatter}



\title{Stochastic optimal well control in subsurface reservoirs using reinforcement learning
\\
\small (Submitted October 28, 2021; Accepted June 17, 2022)}


\author[inst1]{Atish Dixit}

\affiliation[inst1]{organization={School of Energy, Geoscience, Infrastructure and Society},
            addressline={Heriot-Watt University}, 
            city={Edinburgh},
            postcode={EH144AS}, 
            country={UK}}

\author[inst1]{Ahmed H. ElSheikh}


\begin{abstract}
We present a case study of model-free reinforcement learning (RL) framework to solve stochastic optimal control for a predefined parameter uncertainty distribution and partially observable system. We focus on robust optimal well control problem which is a subject of intensive research activities in the field of subsurface reservoir management. 
For this problem, the system is partially observed since the data is only available at well locations. Furthermore, the model parameters are highly uncertain due to sparsity of available field data. 
In principle, RL algorithms are capable of learning optimal action policies — a map from states to actions — to maximize a numerical reward signal. In deep RL, this mapping from state to action is parameterized using a deep neural network.
In the RL formulation of the robust optimal well control problem, the states are represented by saturation and pressure values at well locations while the actions represent the valve openings controlling the flow through wells. 
The numerical reward refers to the total sweep efficiency and the uncertain model parameter is the subsurface permeability field. 
The model parameter uncertainties are handled by introducing a domain randomisation scheme that exploits cluster analysis on its uncertainty distribution.
We present numerical results using two state-of-the-art RL algorithms, proximal policy optimization (PPO) and advantage actor-critic (A2C), on two subsurface flow test cases representing two distinct uncertainty distributions of permeability field. 
The results were benchmarked against optimisation results obtained using differential evolution algorithm. 
Furthermore, we demonstrate the robustness of the proposed use of RL by evaluating the learned control policy on unseen samples drawn from the parameter uncertainty distribution that were not used during the training process. 
\end{abstract}




\end{frontmatter}


\section{\label{sec:level1}Introduction}
Optimal control problem involves finding controls for a dynamical system such that a certain objective function is optimized. 
Traditionally, most research for solving the optimal control problems for non-linear dynamical systems uses optimal control theory which, in principle, finds optimal control by deriving Pontryagin's maximum principle or by solving the Hamilton–Jacobi–Bellman equation.
These classical strategies are offline and require a complete knowledge of system dynamics making them unsuitable for dynamical systems with uncertainties \citep{kiumarsi2017optimal}. 
Recently, model predictive control (MPC) -- a feedback control based on real-time optimisation -- has attracted increasing attention in stochastic optimal control research \citep{diehl2005real, biegler1991optimization, kouvaritakis2001non}. 
Alternatively, optimal control problems could also be solved using reinforcement learning (RL) approaches. \citet{koryakovskiy2017benchmarking} provide a benchmark study for comparison between model predictive control and model-free reinforcement learning approaches where the authors show that RL results are comparable to MPC. 
Further, after a certain break-even point model-free reinforcement learning is shown to perform better than nonlinear model predictive control with an inaccurate model. 
Furthermore, as opposed to MPC, RL provides an extra advantage of generality since it doesn't need complete knowledge of model dynamics. 
So far, research on application of RL in optimal control problems is advancing rapidly in fields like manufacturing \citep{dornheim2020model}, energy \citep{anderlini2016control} and fluid dynamics \citep{rabault2019artificial}.
However, research focused on developing RL strategy based on assessment of its robustness against the uncertainties is still an open area, especially for cases where model uncertainties has a substantial effect on the optimal control. 

In this study, we utilize a model-free RL algorithm to solve simulation-based robust optimal control problem where the model information is assumed to be partially available.
\textit{Robust optimal well control} in petroleum reservoir management \citep{van2009robust, roseta2004robust, brouwer2001recovery} forms a perfect candidate for such problem.
In this problem, the reservoir well control variables are optimized in order to maximize the sweep efficiency of injection fluid throughout the reservoir life cycle.
The assumption of partially available model information is based on the fact that the reservoir field data is generally only available at well locations.
We consider reservoir permeability field as an uncertain model parameter. 
Results are demonstrated for two state of the art model-free RL algorithms: proximal policy optimisation \citep{schulman2017proximal} and advantage actor-critic \citep{mnih2016asynchronous}. 
Although we utilize robust well control in the current study, the presented techniques are general and is applicable to a wide range of simulation-based nonlinear optimal control problems.

We designed two test cases that underscore effect of parameter uncertainty on optimal controls. For ease of execution and demonstration, we represent the optimal well control problem with an advection equation for tracer flow through porous media in which uncertain parameter, permeability, is treated as a random variable. 
In the first test case, we use a permeability field distribution where the location and orientation of a linear high permeability channel is uncertain. 
The second test case uses a spatially correlated permeability field to represent smoother permeability field where the log-permeability field is modeled with a conditional Gaussian distribution with an exponential kernel.
We employed proximal policy optimisation (PPO) and advantage actor-critic (A2C) algorithms to learn the optimal well control policy in order to maximize the sweep efficiency in the domain. 
The RL policies are learned using a selected number of realizations of uncertainty distribution and its robustness is evaluated by applying the learned policy on a set of unseen realizations during training, drawn from the same distribution.

The outline of the rest of this paper is as following: Section \ref{sect: prob_desc} provides the problem description and how RL algorithms are utilized to solve robust optimal well control.
Information such as algorithms and methodologies used in this paper are also briefed in this section. 
Section \ref{sect: num_expt} details the model parameters for test cases chosen for this study. 
Further, the approach used for problem formulation for RL algorithms is explained. Results for the given test cases are presented in section \ref{sect: results}. Finally, section \ref{sect: conclusion} concludes with the research study summary and few future research directions.

\section{Methodology} \label{sect: prob_desc}
The process of single-phase flow in porous media is of importance to a variety of engineers and scientists who are concerned with problems ranging from the financial aspects of oil movement in petroleum reservoirs to the social problems of groundwater
flows in polluted aquifers \citep{whitaker1999single}. 
We demonstrate the proposed techniques in the context of subsurface flow in subsurface reservoirs using an advection equation for tracer flow through porous media.
In this dynamical system, the \textit{tracer} enters the domain and pushes the \textit{non-traced fluid} out of the domain. Flow in and out of the domain is defined as the source and sink terms in the advection equation.
In the context of oil movement problem, the tracer corresponds to water and the non-traced fluid corresponds to oil (or hydrocarbons) in the reservoir while the source and sink locations correspond to injector and producer wells, respectively. 
Bear in mind that the oil flow problem can be more correctly modeled as a two-phase problem (water phase and oil phase). However for ease of execution and demonstration we chose single-phase flow problems where water injection is modeled as tracer flow that pushes the oil (non-traced fluid) in the reservoir, much like a contaminant in a fluid. 
Despite this approximation, the presented methodology is general enough and can be similarly applied to two-phase flow problems.

\subsection{Problem description for robust optimal well control}
We model the reservoir water injection process with an advection equation for tracer flow through porous media over the temporal domain $\mathcal{T} = [t_0, T] \subset \mathbb{R}$ and spatial domain $\mathcal{X} \subset \mathbb{R}^2$. 
The tracer flow models water flooding with the fractional variable $s(x,t) \in [0,1]$, where $s(x,t)$ represents the fraction of water to oil at $x \in \mathcal{X}$ and $t \in \mathcal{T}$. 
The well controls $a(x',t)$ represent the source and sink terms in this equation, where $x' \in \mathcal{X'} $ (where $\mathcal{X'} \subset \mathcal{X}$) correspond to set of well locations. 
Injector and producer flow controls are represented with $a^+$ (which constitutes of all the source locations in the domain and formulated as $\max(0,a)$) and $a^-$ (which constitutes of all the sink locations in the domain and formulated as $\min(0,a)$), respectively. 
The optimal controls $a^*(x',t)$ are the solution of following closed-loop optimisation problem:
\begin{subequations}
\begin{align}
    & \max_{s(\cdot), a(\cdot)} \int_{t_0}^T  \left ( \sum_{x'}a^-(x',t)(1-s(x',t))  \right )  dt, & x' \in \mathcal{X'}, \ t \in \mathcal{T} \label{eq: obj_fun} \\
    & \frac{ds}{dt} = \frac{1}{\phi} \left (a^+ + sa^- - \nabla \cdot sv \right ), & x \in \mathcal{X}, \ t \in \mathcal{T} \label{eq: gov_eq} \\
    & s(\cdot,0) = s_0,\ \ v \cdot \textbf{n} = 0, & \label{eq: init_eq}\\
    & \sum_{x'} a^+(x',t)  = -\sum_{x'} a^-(x',t) = c, & x' \in \mathcal{X'}, \ t \in \mathcal{T} \label{eq: constr}
\end{align}
\label{eq: prob_def}
\end{subequations}
where, the objective function defined in Eq.~\eqref{eq: obj_fun} represents the total oil flow out of the reservoir (oil production) and is maximized over the finite time interval $\mathcal{T}$. 
The intigrand in this function is referred as Lagrangian term in control theory and is often denoted by $L(s,a)$. 
The water flow trajectory $s(x,t)$ is governed by an advection equation (Eq.~\eqref{eq: gov_eq}) which is solved in couple with pressure equation $-\nabla \cdot (k/\mu) \nabla p = a $, where $p(x,t) \in \mathbb{R}$ is pressure. 
Porosity $\phi (x,\cdot)$, permeability $k (x,\cdot)$ and viscosity $\mu (x,\cdot)$ are the model parameters.
Permeability $k$ represents the model uncertainty and is treated as a random variable that follows a known probability density function $\mathcal{K}$ with $K$ as its domain. 
The initial and no flow boundary condition is defined in Eq.~\eqref{eq: init_eq}, where \textbf{n} denotes outward normal vector from the boundary of $\mathcal{X}$. 
Note that, the velocity $v$ follows Darcy's law: $v = - (k/\mu) \nabla p$. 
Constraint defined in Eq.~\eqref{eq: constr} represents the fluid incompressibility assumption along with the fixed total source/sink term $c$ which represents water injection rate in the reservoir.
The controls $a(x',t)$ are discretized on time interval $t_0<t_1<\cdots <t_{m}<t_{m+1}=T$. 

\subsection{RL framework for robust optimal well control}
We propose to use model-free reinforcement learning algorithms to solve the optimal control problem defined in Eq.~\eqref{eq: prob_def}. 
A common approach in RL is to model the task as a Markov decision process. 
The process is defined as a quadruple $\left \langle \mathcal{S},\mathcal{A},\mathcal{P},\mathcal{R} \right \rangle$, where $\mathcal{S} \subset \mathbb{R}^{n_s}$ is set of all possible states with dimension $n_s$, $\mathcal{A} \subset \mathbb{R}^{n_a}$ is a set of all possible actions with dimension $n_a$. 
State is represented with the tracer variable $s(x,t_m)$ and action with controls $a(x',t_m)$ such that it follows the constraint defined in Eq.~\eqref{eq: constr}. 
$\mathcal{P}:\mathcal{S}\times \mathcal{A} \rightarrow \mathcal{S}$ is a transition function that follows \textit{Markov property}. 
That is, it returns $s(x,t_{m+1})$ as a function of control $a(x',t_{m})$ and state $s(x,t_{m})$. 
Such transition function can be obtained by discretizing Eq.~\eqref{eq: gov_eq} which returns $s(x,t_{m+1})$ by executing the controls $a(x',t_{m})$ when in the state $s(x,t_{m})$.
The reward function $\mathcal{R}:\mathcal{S} \times \mathcal{A} \times \mathcal{S} \rightarrow \mathbb{R}$ returns a real valued reward, $r(t_{m+1})=\mathcal{R}(s(\cdot,t_{m}), a(\cdot,t_{m}), s(\cdot,t_{m+1}))$ for a particular transition between the states. 
The reward function for the problem (Eq.~\eqref{eq: prob_def}) is obtained by discretizing the objective function (Eq.~\eqref{eq: obj_fun}) into control steps such that,
\begin{equation}
    r(t_{m+1})= \int_{t_m}^{t_{m+1}}  L(s,a) dt.
    \label{eq: reward_func}
\end{equation}
The control policy, $\pi:\mathcal{S} \rightarrow \mathcal{A}$, defines an action $a(x',t_m)$ given the current state $s(x,t_m)$ (also written as $\pi(a|s)$).

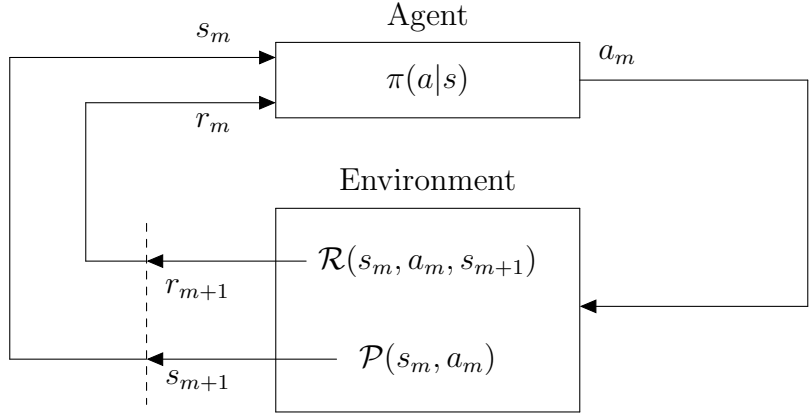
\begin{figure}[ht]
    \centering
    \begin{tikzpicture}[line cap=round,line join=round,>=triangle 45,x=1cm,y=1cm, scale=1.0]
\clip(-6,-3) rectangle (6,3);

\draw [line width=0.4pt] (-2,2)-- (-2,1);
\draw [line width=0.4pt] (-2,1)-- (2,1);
\draw [line width=0.4pt] (2,1)-- (2,2);
\draw [line width=0.4pt] (2,2)-- (-2,2);

\draw [line width=0.4pt] (-2,-0.2)-- (-2,-2.9);
\draw [line width=0.4pt] (-2,-2.9)-- (2,-2.9);
\draw [line width=0.4pt] (2,-2.9)-- (2,-0.2);
\draw [line width=0.4pt] (2,-0.2)-- (-2,-0.2);

\draw [->,line width=0.4pt] (-1.6,-0.9) -- (-3.7,-0.9);
\draw [->,line width=0.4pt] (-1.2,-2.2) -- (-3.7,-2.2);
\draw [line width=0.4pt] (-4.5,-0.9) -- (-3.7,-0.9);
\draw [line width=0.4pt] (-5.5,-2.2) -- (-3.7,-2.2);
\draw [dashed] (-3.7, -0.4) -- (-3.7, -2.8);
\draw [->,line width=0.4pt] (-4.5,1.2) -- (-2,1.2);
\draw [->,line width=0.4pt] (-5.5,1.8) -- (-2,1.8);
\draw [line width=0.4pt] (-4.5,-0.9)-- (-4.5,1.2);
\draw [line width=0.4pt] (-5.5,-2.2)-- (-5.5,1.8);
\draw [line width=0.4pt] (2,1.5) -- (5,1.5);
\draw [->,line width=0.4pt] (5,-1.5) -- (2,-1.5);
\draw [line width=0.4pt] (5,1.5)-- (5,-1.5);

\draw (0,1.5) node[anchor=center] {$\pi(a|s)$};
\draw (0,-2.2) node[anchor=center] {$\mathcal{P}(s_m, a_m)$};
\draw (0,-0.9) node[anchor=center] {$\mathcal{R}(s_m, a_m, s_{m+1})$};
\draw (2.1,2.1) node[anchor=north west] {$a_m$};
\draw (-3.6,-1) node[anchor=north west] {$r_{m+1}$};
\draw (-3.6,-2.2) node[anchor=north west] {$s_{m+1}$};
\draw (-3.2,1.2) node[anchor=north west] {$r_m$};
\draw (-3.2,2.4) node[anchor=north west] {$s_m$};
\draw (0,2) node[anchor=south] {$\mathrm{Agent}$};
\draw (0,-0.1) node[anchor=south] {$\mathrm{Environment}$};
\end{tikzpicture}
    \caption{A typical agent-environment interaction in RL algorithms. state $s(x,t_m)$, action $a(x',t_m)$ and reward $r(t_m)$ are denoted with shorthand notations, $s_m$, $a_m$ and $r_m$, respectively}
    \label{fig: rl_framework}
\end{figure}

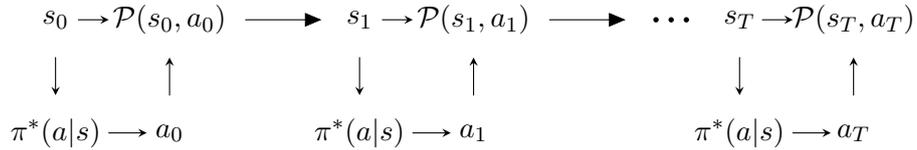
\begin{figure}[ht]
    \centering
    \begin{tikzpicture}[line cap=round,line join=round,>=triangle 45,x=1cm,y=1cm]
\clip(-6.1,-2) rectangle (6,0.5);

\draw [->,>=stealth,line width=0.4pt] (-5.2,0)-- (-4.8,0);
\draw [->,>=stealth,line width=0.4pt] (-1.2,0)-- (-0.8,0);
\draw [->,>=stealth,line width=0.4pt] (3.8,0)-- (4.2,0);
\draw [->,line width=0.4pt] (-3,0)-- (-2,0);
\draw [->,line width=0.4pt] (1,0)-- (2,0);

\draw [->,>=stealth,line width=0.4pt] (-4.8,-1.5) -- (-4.3,-1.5);
\draw [->,>=stealth,line width=0.4pt] (-0.8,-1.5) -- (-0.3,-1.5);
\draw [->,>=stealth,line width=0.4pt] (4.2,-1.5) -- (4.7,-1.5);

\draw [->,>=stealth,line width=0.4pt] (-5.5,-0.5)--(-5.5,-1);
\draw [<-,>=stealth,line width=0.4pt] (-4,-0.5)-- (-4,-1);

\draw [->,>=stealth,line width=0.4pt] (-1.5,-0.5)--(-1.5,-1);
\draw [<-,>=stealth,line width=0.4pt] (0,-0.5)--(0,-1);

\draw [->,>=stealth,line width=0.4pt] (3.5,-0.5)--(3.5,-1);
\draw [<-,>=stealth,line width=0.4pt] (5,-0.5)--(5,-1);

\draw [fill=black] (2.6,0) circle (0.8pt);
\draw [fill=black] (2.8,0) circle (0.8pt);
\draw [fill=black] (2.4,0) circle (0.8pt);

\draw (-5.5,0) node[anchor=center] {\small $s_0$};
\draw (-4,0) node[anchor=center] {\small $\mathcal{P}(s_0, a_0)$};
\draw (-1.5,0) node[anchor=center] {\small $s_1$};
\draw (0,0) node[anchor=center] {\small $\mathcal{P}(s_1, a_1)$};
\draw (3.5,0) node[anchor=center] {\small $s_T$};
\draw (5,0) node[anchor=center] {\small $\mathcal{P}(s_T, a_T)$};

\draw (-5.5,-1.5) node[anchor=center] {\small $\pi^*(a|s)$};
\draw (-4,-1.5) node[anchor=center] {\small $a_0$};
\draw (-1.5,-1.5) node[anchor=center] {\small $\pi^*(a|s)$};
\draw (0,-1.5) node[anchor=center] {\small $a_1$};
\draw (3.5,-1.5) node[anchor=center] {\small $\pi^*(a|s)$};
\draw (5,-1.5) node[anchor=center] {\small $a_T$};

\end{tikzpicture}
    \caption{optimal controls for complete control trajectory which refers to an episode in RL algorithms. state $s(x, t_m)$ and action $a(x',t_m)$ are denoted with shorthand notations, $s_m$ and $a_m$, respectively}
    \label{fig: rl_episode}
\end{figure}

The goal of reinforcement learning is to find an optimal policy $\pi^*(a|s)$ that maximizes expected discounted return, $\textbf{G} = \sum_{m=1}^M \gamma^{m-1} r(t_{m})$, where immediate rewards $r$ are exponentially decayed by the discount rate $\gamma \in [0,1]$ and $M$ is the final control time step. 
Essentially, RL algorithms attempt to learn the optimal policy $\pi^*(a|s)$ from an initial policy, $\pi(a|s)$, by exploring state-action space with what is referred to as agent-environment interactions. Figure \ref{fig: rl_framework} shows a typical schematic of such agent-environment interaction. The term \textit{agent} refers to the controller that follows the policy $\pi(a|s)$ while the \textit{environment} consists of transition function, $\mathcal{P}$, and reward function, $\mathcal{R}$.
The optimum solution can be obtained by following the policy $\pi^*(a|s)$ throughout the complete control trajectory (also referred as an \textit{episode} in RL literature) as shown in figure \ref{fig: rl_episode}.
Thus, optimal result (which refers to optimum oil recovery for optimal well control problem), $\textbf{R}^{\pi^*(a|s)}$, is obtained by adding the reward $r(\cdot)$ at each time-step of such episode:
\begin{equation}
    \textbf{R}^{\pi^*(a|s)} = \sum_{m=0}^{M-1}  \left [ \int_{t_m}^{t_{m+1}} L(s, \pi^*(a|s))dt \right ].
    \label{eq: opt_res}
\end{equation}
Note that the optimal result, which is used in the rest of the paper to evaluate the policy, is not exponentially decayed with the discount factor $\gamma$.
The policy $\pi^*(a|s)$ is said to be \textit{robust} if it is able to achieve optimal results for a stochastic environment controlled by the uncertain parameter $k$, defining the permeability in the Darcy flow equation.

If we treat the parameter $k$ as a deterministic fixed value, policy learning is fairly straightforward. The policy learned in such a way is termed as \textit{frozen} policy (a term used by \cite{koryakovskiy2017benchmarking}). 
In this study we aim to find a robust policy that accounts for the variability in $k$. 
To the best of our knowledge, so far, such policy is learned by simply incurring samples from the distribution $\mathcal{K}$, in each episode of the learning process (also known as static domain randomisation method \cite{muratore2021robot}).
Such policy can be robust enough if the samples used in the learning process cover most of its domain $K$ (In other words, when $K$ is well explored). 
For this reason, robust optimal policy learning naturally requires higher number of agent-environment interactions as compared to that in learning frozen policy. 
This could be problematic if each agent-environment interaction (solving the governing Equations \eqref{eq: gov_eq}, for instance) is computationally intensive, which is common in most simulation-based optimal control problems like optimal well control. 
Furthermore, samples incurred in this learning process often tend to be from the high probability region of the distribution domain causing the policy to be biased towards them.

\textit{robustness of frozen policy}: We denote frozen policy learned by keeping the parameter $k$ fixed as $\pi^*(a|s;k)$.
Lets define a distance function $D: K \times K \rightarrow \mathbb{R}$ which returns the distance between $k$ and $k'$ as $D(k, k')$. 
Naturally, the frozen policy can be applied to the simulation (or environment) that uses $k'$ as the parameter instead of $k$ (denoted as $\pi^*(a|s;k\Rightarrow k')$). 
Due to continuous nature of governing Equations \eqref{eq: gov_eq}, effectiveness of the policy, $\pi^*(a|s;k\Rightarrow k')$, is inversely related to the distance $D(k, k')$. 
We can define an acceptable near-optimal solution limit obtained with $\pi^*(a|s;k\Rightarrow k')$ when $k'$ is in the neighborhood ($\delta$) of $k$. 
In other words, we can say the policy $\pi^*(a|s;k\Rightarrow k')$ can be considered robust when $D(k,k')<\delta$.

\begin{figure}
    \centering
    \input{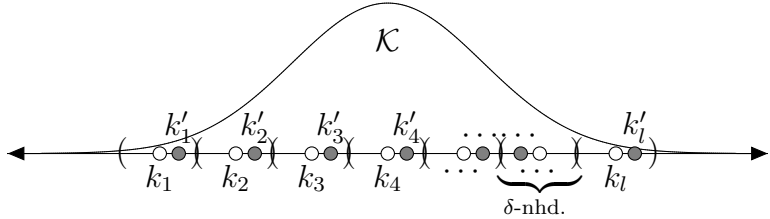}
    \caption{a wells spread choice of samples for some uncertainty distribution $\mathcal{K}$ (depicted as one-dimensional for illustration purpose) to learn the robust optimal policy $\pi^*(a|s; \textbf{k})$. Members of $\textbf{k}$ are colored in white while the members of $\textbf{k}'$ are shown in 
    grey.}
    \label{fig: k_samples}
\end{figure}

This argument can be extended for multiple parameters: $k_1, k_2, \cdots, k_l$. In this case, we learn the policy, $\pi^*(a|s; \textbf{k})$, by randomly choosing any one of the parameter value from the vector $\textbf{k}=( k_1, k_2, \cdots, k_l )$ at every episode in the learning process.
Subsequently, the policy, $\pi^*(a|s; \textbf{k} \Rightarrow \textbf{k}')$, can yield a near optimal value for the vector $\textbf{k}'=( k'_1, k'_2, \cdots, k'_l )$, given that $\min_{i} D(k_i, k'_j) < \delta$, $\forall j \in \{1,2,\cdots, l\}$.
If the vector $\textbf{k}$ is chosen such that the union of neighbourhood of all its values $k_i,\ \forall i \in \{1,2,\cdots, l\},$ cover the domain $K$, the policy $\pi^*(a|s; \textbf{k})$ yields at least near optimal solution for any sample $k' \sim \mathcal{K}$. Figure \ref{fig: k_samples} shows an example of such choice of $\textbf{k}$ values for the uncertainty distribution $\mathcal{K}$.
For such a \textit{well spread} choice of $\textbf{k}$, the policy $\pi^*(a|s;\textbf{k})$ can be said to be robust under the uncertainty distribution $\mathcal{K}$.

\textbf{Implementation:} Although the above-mentioned policy $\pi^*(a|s)$ provides optimal solution for the problem defined in Equation~\eqref{eq: prob_def}, it is not applicable for systems with partially observable state space. 
For instance, in optimal well control problems reservoir information is generally only available at well locations. 
As a result we provide the agent with the available \textit{observation} $o(x',t_m)$ as its state. For this study, observation $o(x',t_m)$ is represented with a set of saturation and pressure values at well locations $x'$ and time $t_m$. 
Note that, with such representation of states, we break the underlying assumption of Markov property of the transition function. 
Here, we assume that transition between the observations approximately follow the Markov property. 

We choose $l$ well spread samples of uncertainty distribution to learn a robust policy. 
This is achieved with a clustering analysis  (using k-means clustering method) of the domain $K$. 
The \textit{training vector} $\textbf{k}$ is constructed with samples of $\mathcal{K}$ which are located at the cluster centers. 
The policy, $\pi^*(a|s;\textbf{k})$, is learned by randomly selecting the parameter $k$ from the training vector $\textbf{k} = \{k_1, k_2, \cdots k_l\}$. 
Average \textit{training return} is calculated by averaging the returns of policy $\pi(a|s;\textbf{k})$ on $l$ simulations, each with a separate parameter $k$ from \textbf{k}. 
\begin{equation}
    \textbf{R}^{\pi(a|s; \textbf{k} \Rightarrow \textbf{k})} = \frac{1}{l}  \sum_{i=1}^l \left [ \sum_{m=0}^{M-1} \left ( \int_{t_m}^{t_{m+1}} L(s, \pi(a|s; \textbf{k} \Rightarrow k_i)) dt  \right ) \right ]
\end{equation}
The robustness of this policy is assessed by applying it on $l$ simulations, each with a new unseen sample, $k' \sim \mathcal{K}$, as its parameter.
The samples for evaluation are chosen randomly from each cluster.
The \textit{evaluation vector} $\textbf{k}'=\{k'_1, k'_2, \cdots k'_l \}$ represents the set of such samples.
Robustness of the policy $\pi(a|s;\textbf{k})$ is evaluated by monitoring the average \textit{evaluation return} $\textbf{R}^{\pi(a|s; \textbf{k} \Rightarrow \textbf{k}')}$:
\begin{equation}
    \textbf{R}^{\pi(a|s; \textbf{k} \Rightarrow \textbf{k}')} = \frac{1}{l}  \sum_{i=1}^l \left [ \sum_{m=0}^{M-1} \left ( \int_{t_m}^{t_{m+1}} L(s, \pi(a|s; \textbf{k} \Rightarrow k'_i)) dt  \right ) \right ].
\end{equation}
In a nutshell, we  explect to train the policy $\pi(a|s; \textbf{k})$ such that $\textbf{R}^{\pi(a|s; \textbf{k} \Rightarrow \textbf{k}')}$ is maximized. 

We employ PPO and A2C algorithms to solve this problem.
The RL algorithm parameters are tuned in order to maximize the average returns,  $\textbf{R}^{\pi(a|s; \textbf{k} \Rightarrow \textbf{k})}$. 
As the values of training vector, \textbf{k}, fairly represent the variety of the domain $K$, we expect convergence of average \textit{evaluation} return value towards $\textbf{R}^{\pi^*(a|s; \textbf{k} \Rightarrow \textbf{k}')}$ by the end of the learning process. 
The baseline optimal results are computed using differential evolution (DE) algorithm \citep{storn1997differential}. 
DE algorithm is used to solve $l$ optimisation problems as described in Equations \eqref{eq: prob_def} each with the parameter $k$ replaced by a fixed value from evaluation vector $\textbf{k}'$.
The reference value for $\textbf{R}^{\pi^*(a|s; \textbf{k} \Rightarrow \textbf{k}')}$ is taken as average of these $l$ solved optimal objective functions. 
Both the RL algorithms are tuned such that $\textbf{R}^{\pi(a|s; \textbf{k} \Rightarrow \textbf{k}')}$ converges in the range of this reference value.

\subsection{RL algorithms}

We choose model-free RL algorithms to avoid any effect of model learning on policy learning.
Two state-of-the-art policy-based algorithms (A2C and PPO) are used to solve the optimal control problem under consideration.  

\subsubsection{Advantage actor-critic algorithm}

A2C \citep{mnih2016asynchronous} is a policy gradient algorithm that models the stochastic policy, $\pi_\theta(s|a)$, with a neural network. Essentially, the network parameters $\theta$ are obtained by optimizing for the objective function,
\begin{equation}
    J_{a2c}(\theta) = \hat{\mathbb{E}}_t\left [ \log\pi_{\theta}(a_t|s_t) \hat{A}_t \right ],
    \label{eq: a2c_grad}
\end{equation}
where, $\hat{A}_t$ is an estimator for advantage function at timestep $t$ and the term, $\hat{\mathbb{E}}_t[\cdots]$, is empirical average over finite batch of samples collected through agent-environment interactions. 
Gradient estimator of policy network, $\nabla_{\theta}J(\theta)$, is obtained by differentiating Eq.~\eqref{eq: a2c_grad} which is done with automatic differentiation algorithm. The advantage function estimator, $\hat{A}_t$, is computed using generalized advantage estimator \citep{schulman2015high} which is derived from the value function $V_t$. 
The value function estimator $\hat{V}_t$ is learned through a separate neural network termed as the $critic$ network. Definitions of advantage and value functions are described in \ref{app: value_funs}. 
Algorithm \ref{alg:a2c} illustrates a broad outline for implementation of A2C algorithm in this study. 
In order to reduce computational time, the iterative data sampling for objective function is performed in parallel on $N$ processors followed by a synchronous gradient update. 
Note that in every policy iteration in total $T$ control steps are run where environment is reset with a new permeability sample from the $\textbf{k}$ after every terminal step of the episode.

\begin{algorithm}
	\caption{Policy Robustness Evaluation using A2C} 
	\begin{algorithmic}[1]
	    \State \textbf{Input}: Number of actors $N$, and number of steps in each policy iteration $T$
	    \State Obtain training vector $\textbf{k}$, and evaluation vector $\textbf{k}'$ using cluster analysis of the predefined uncertainty distribution $\mathcal{K}$
		\For {$iteration=1,2,\ldots$}
			\For {$actor=1,2,\ldots,N$}
				\State \multiline{ Run policy $\pi_{\theta}$ in environment for $T$ time steps (which corresponds to in total $E$ episodes where the environment permeability  is set to a sample from training vector $\textbf{k}$, at the beginning of every episode ) }
				\State Compute value function estimates $\hat{V}_1,\ldots,\hat{V}_T$ using critic network
				\State Compute advantage function estimates $\hat{A}_1,\ldots,\hat{A}_T$ 
			\EndFor
			\State Optimize $J_{a2c}(\theta)$\footnotemark (Eq.~\eqref{eq: a2c_grad}), with single epoch and batch size $NT$
			\State $\theta_{old}\leftarrow\theta$
			\State \multiline{ compute and record training return $\textbf{R}^{\pi(a|s; \textbf{k} \Rightarrow \textbf{k})}$, and evaluation return $\textbf{R}^{\pi(a|s; \textbf{k} \Rightarrow \textbf{k}')}$}
		\EndFor
	\end{algorithmic} 
	\label{alg:a2c}
\end{algorithm}
\footnotetext{In practice, a single integrated neural network is used for both, actor and critic networks. As a result, objective function for automatic differentiation is summation of Eq.~\eqref{eq: a2c_grad} and value loss function for critic network. Please refer Algorithm S2 from \cite{mnih2016asynchronous} for the formulation of value loss function. }

\subsubsection{Proximal policy optimisation algorithm} \label{sect:ppo}
 If the gradient step in A2C is too large, the policy may astray which in turn will produce bad samples causing divergence in the solution. 
 As a result, we have to select very small step size which slows the learning process. \citet{schulman2017proximal} introduced PPO algorithm that make sure the gradient steps are small enough to make the algorithm data efficient. 
 This is done by formulating the network objective function in terms of a ratio of two policies (old and new) using principle of importance sampling.
 Appropriate gradient steps are chosen by clipping the ratio of old and new policy within the range $[1-\epsilon,1+\epsilon]$, where $\epsilon$ is generally a small fractional number. 
 The modified objective function for policy network is defined as,
\begin{equation}
    J_{ppo}(\theta) = \hat{\mathbb{E}}_t  \left [ \min( r_t(\theta) \hat{A}_t (s_t,a_t),\  \textup{clip}(r_t(\theta),1-\epsilon,1+\epsilon)\hat{A}_t (s_t,a_t) ) \right ],
    \label{eq: ppo_loss}
\end{equation}
where $r_t(\theta) = \pi_{\theta}(a_t|s_t)/\pi_{\theta_{old}}(a_t|s_t)$ and $\pi_{\theta_{old}}(a_t|s_t)$ is old policy. Algorithm \ref{alg:ppo} illustrates the implementation of PPO algorithm in the presented methodology.  

\begin{algorithm}
	\caption{Policy Robustness Evaluation using PPO} 
	\begin{algorithmic}[1]
	\State \textbf{Input}: Number of actors $N$, number of steps in each policy iteration $T$, number of epochs $K$ and minibatch size $M$
	\State Obtain training vector $\textbf{k}$, and evaluation vector $\textbf{k}'$ using cluster analysis of the predefined uncertainty distribution $\mathcal{K}$
		\For {$iteration=1,2,\ldots$}
			\For {$actor=1,2,\ldots,N$}
				\State \multiline{ Run policy $\pi_{\theta}$ in environment for $T$ time steps (which corresponds to in total $E$ episodes where the environment permeability  is set to a sample from training vector $\textbf{k}$, at the beginning of every episode ) }
				\State Compute value function estimates $\hat{V}_1,\ldots,\hat{V}_T$ using critic network
				\State Compute advantage function estimates $\hat{A}_1,\ldots,\hat{A}_T$ 
			\EndFor
			\State Optimize $J_{ppo}(\theta)$\footnotemark (Eq.~\eqref{eq: ppo_loss}), with $K$ epochs and minibatch size $M\leq NT$
			\State $\theta_{old}\leftarrow\theta$
			\State \multiline{ compute and record training return $\textbf{R}^{\pi(a|s; \textbf{k} \Rightarrow \textbf{k})}$, and evaluation return $\textbf{R}^{\pi(a|s; \textbf{k} \Rightarrow \textbf{k}')}$}
		\EndFor
	\end{algorithmic} 
	\label{alg:ppo}
\end{algorithm}
\footnotetext{ The objective function for integrated actor-critic network, in the PPO implementation, is the summation of actor loss term (Eq.~\eqref{eq: ppo_loss}), value loss term and entropy loss term. Readers are referred to \cite{schulman2017proximal} for the detailed definition of value loss term and entropy loss term. }

\subsection{Differential evolution algorithm} \label{ref: deff_ev}
We employ DE algorithm \citep{storn1997differential} as a baseline for assessing the results obtained using PPO and A2C. 
Essentially, DE is a population based stochastic optimisation algorithm which employs evolutionary ideas like crossover and mutation to find optimal arguments.
In order to solve the optimisation problem (Eq.~\eqref{eq: prob_def}) using DE, we represent the argument with a set of controls $\textbf{a} = \{ a(x',t_0), \cdots , a(x',T) \}$ such that it follows the constraints defined in Eq.~\eqref{eq: constr}. 
The \textit{fitness} of such argument is computed with the objective function (Eq.~\eqref{eq: obj_fun}) by solving the governing Eq.~\eqref{eq: gov_eq}. 
DE algorithm initiates its argument search with a set $\textbf{A}$ of random arguments which is referred as $\textit{population}$ (i.e. $\textbf{A}=\{ \textbf{a}_{1},\cdots, \textbf{a}_{p} \}$). 
Using a crossover criteria, certain arguments (say, $i^{th}$ argument in $\textbf{A}$: $\textbf{a}_i$) are evolved as ,
\begin{equation*}
    \textbf{a}'_i = \textbf{a}^* + F(\textbf{a}_{r1} + \textbf{a}_{r2}),
\end{equation*}
where $\textbf{a}'_i$ is updated value for $\textbf{a}_i$, $\textbf{a}^*$ is the best argument (i.e. the one corresponding to maximum fitness) in the population so far, $F \in [0,2]$ is mutation parameter, $\textbf{a}_{r1}$ and $\textbf{a}_{r2}$ are randomly selected arguments from the population.
$\textbf{a}_i$ is replaced with $\textbf{a}'_i$ if the fitness of $\textbf{a}'_i$ is higher than that for $\textbf{a}_i$. The optimum solution is obtained by repeating such evolution for a number of iterations or until a certain convergence criteria is met.

Note that DE algorithm's parameter search space is wider than that for RL. This is because the optimum parameters do not have to follow a mapping $\pi(a|s)$.
For this reason, we expect DE algorithm to achieve more optimal controls as compared to RL algorithms due to its potential to achieve global optima. 

\subsection{K-means clustering} \label{sect: kmeans}
We employ connectivity distance \citep{caers2009modeling} measure in order to represent the variation in dynamical response of permeability samples. 
The connectivity distance matrix $\textbf{D} \in \mathbb{R}^{N \times N}$ for a large number ($N$) of samples of $\mathcal{K}$ is written by,
\begin{equation}
    \textbf{D}(k_i, k_j) = \sum_{x''} \int_{t_0}^{T} \left [ s(x'',t; k_i) - s(x'',t; k_j) \right ]^2 dt,
    \label{eq: conn_dist}
\end{equation}
where, $x''$ are a set of spatial locations in the domain $\mathcal{X}$ and $s(x'', t; k_i)$ refers to solution of governing Eq.~\eqref{eq: gov_eq} with the uncertainty parameter $k_i$ when all control wells are kept equally open. 
In order to be able to visualize the connectivity dissimilarity among samples of $\mathcal{K}$, we employ multi-dimensional scaling on the distance matrix $\textbf{D}$ to obtain a set of $N$ two-dimensional coordinates represented with $d_1, d_2, \cdots , d_N$.
In other words, coordinates $d_1, d_2, \cdots , d_N$ correspond to samples $k_1, k_2, \cdots , k_N$ of $\mathcal{K}$ such that it represents connectivity distance measure defined in Eq.~\eqref{eq: conn_dist} among its values.
The coordinates $d_1, d_2, \cdots , d_N$ are divided in $l$ sets $S_1, S_2, \cdots , S_l$ obtained by solving the optimisation problem:
\begin{equation*}
    \arg \min_{S} \sum_{i}^{l} \sum_{d_j \in S_i } \left \|   d_j - \mu_{S_i} \right \|,
\end{equation*}
where $\mu_{S_i}$ is average of all coordinates in the set $S_i$. The training vector \textbf{k} is a set of $l$ samples of $\mathcal{K}$ where each of its value $k_i$ correspond to the one nearest to $\mu_{S_i}$.

\section{Numerical experiments} \label{sect: num_expt}
We evaluate the effectiveness of RL in solving robust optimal well control problems using two test cases representing two distinct permeability uncertainty distributions. 
Numerical solutions of the governing equations are obtained by using finite volume discretization. The pressure equation is discretized using two point flux approximation method while the saturation equation is discretized using implicit upwind scheme. Readers are referred to \cite{aarnes2007introduction} for more details on numerical methodology. 
For both cases, the values for model parameters emulate those in the benchmark reservoir simulation case, SPE-10 model 2 \citep{christie2001tenth}. 

\subsection{model parameters}
\label{sect: sim_par}
Reservoir simulation parameters for both the cases, corresponding to Eq.~\eqref{eq: prob_def}, are delineated in table \ref{tab:res_model}.
As per the convention in geostatistics, we assume that the distribution of $\log{(k)}$ is known and is denoted by $\mathcal{G}$. 
As a result, we treat $g=\log(k)$ as a random variable in the problem description defined in Eq.~\eqref{eq: prob_def}.
Uncertainty distributions for test cases 1 and 2 are denoted with $\mathcal{G}_1$ and $\mathcal{G}_2$, respectively.

\begin{table}[ht]
    \caption{Reservoir model parameters}
    \centering
    \begin{tabular}{l l l l}
        \hline
         & Test case 1 & Test case 2 & units\\
        \hline
        spatial domain $\mathcal{X}$ & (1200$\times$1200) & (1200$\times$1200) & ft$^2$\\
        temporal domain $\mathcal{T}$ & [0,125] & [0,25] & days \\
        initial saturation $s_0$ & 0.0 & 0.0 & -- \\
        viscosity $\mu$ & 0.3 & 0.3 & cP \\
        porosity $\phi$ & 0.2 & 0.2 & -- \\
        number of producers $n_p$ & 31 & 4 & -- \\
        number of injectors $n_i$ & 31 & 1 & -- \\
        total injector flow $\sum a^+$  & 2304 & 8064 & ft$^2$/day\\
        \hline
    \end{tabular}
    \label{tab:res_model}
\end{table}

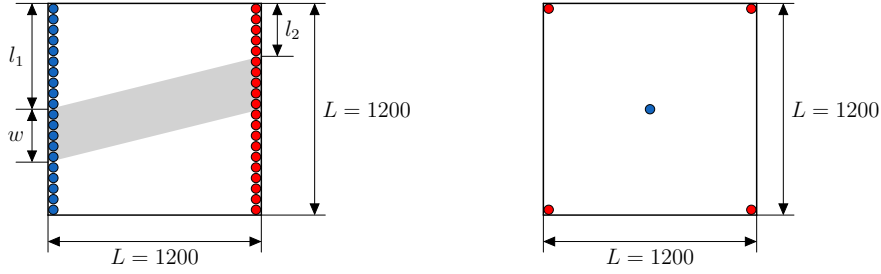
\begin{figure}
    \centering
    \begin{tabular}{c c}
        \subfloat[schematics for test case 1 domain] {\resizebox{0.4\columnwidth}{!} {
        \definecolor{red}{rgb}{1,0,0}
\definecolor{blue}{rgb}{0.08,0.4,0.75}
\begin{tikzpicture}[line cap=round,line join=round,>=triangle 45,x=1cm,y=1cm]
\clip(-3,-3) rectangle (5,3);

\fill[line width=0.8pt,fill=black,fill opacity=0.18] (-2,0) -- (2,1) -- (2,0) -- (-2,-1) -- cycle;

\draw [line width=0.8pt] (2,2)-- (2,-2);
\draw [line width=0.8pt] (2,-2)-- (-2,-2);
\draw [line width=0.8pt] (-2,-2)-- (-2,2);
\draw [line width=0.8pt] (-2,2)-- (2,2);

\draw [line width=0.4pt] (-2,0)-- (-2.6,0);
\draw [line width=0.4pt] (-2,-1)-- (-2.6,-1);
\draw [line width=0.4pt] (-2,2)-- (-2.6,2);
\draw [line width=0.4pt] (2,1)-- (2.6,1);
\draw [line width=0.4pt] (2,2)-- (3.2,2);
\draw [line width=0.4pt] (2,-2)-- (3.2,-2);
\draw [line width=0.4pt] (-2,-2)-- (-2,-2.7);
\draw [line width=0.4pt] (2,-2)-- (2,-2.7);

\draw [<->,line width=0.4pt] (-2.3,0) -- (-2.3,2);
\draw [<->,line width=0.4pt] (-2.3,0) -- (-2.3,-1);
\draw [<->,line width=0.4pt] (2.3,1) -- (2.3,2);
\draw [<->,line width=0.4pt] (3,-2) -- (3,2);
\draw [<->,line width=0.4pt] (-2,-2.5) -- (2,-2.5);

\draw (-2.3,1) node[anchor=east] {$l_1$};
\draw (-2.3,-0.5) node[anchor=east] {$w$};
\draw (2.3,1.5) node[anchor=west] {$l_2$};
\draw (3,0) node[anchor=west] {$L=1200$};
\draw (0,-2.5) node[anchor=north] {$L=1200$};

\draw [fill=blue] (-1.9,1.9) circle (2.5pt);
\draw [fill=blue] (-1.9,1.7) circle (2.5pt);
\draw [fill=blue] (-1.9,1.5) circle (2.5pt);  
\draw [fill=blue] (-1.9,1.3) circle (2.5pt);
\draw [fill=blue] (-1.9,1.1) circle (2.5pt);
\draw [fill=blue] (-1.9,0.9) circle (2.5pt);
\draw [fill=blue] (-1.9,0.7) circle (2.5pt);
\draw [fill=blue] (-1.9,0.3) circle (2.5pt);
\draw [fill=blue] (-1.9,0.1) circle (2.5pt);
\draw [fill=blue] (-1.9,0.5) circle (2.5pt);
\draw [fill=blue] (-1.9,-1.9) circle (2.5pt);
\draw [fill=blue] (-1.9,-1.7) circle (2.5pt);
\draw [fill=blue] (-1.9,-1.5) circle (2.5pt);
\draw [fill=blue] (-1.9,-1.3) circle (2.5pt);
\draw [fill=blue] (-1.9,-1.1) circle (2.5pt);
\draw [fill=blue] (-1.9,-0.9) circle (2.5pt);
\draw [fill=blue] (-1.9,-0.7) circle (2.5pt);
\draw [fill=blue] (-1.9,-0.3) circle (2.5pt);
\draw [fill=blue] (-1.9,-0.1) circle (2.5pt);
\draw [fill=blue] (-1.9,-0.5) circle (2.5pt);
\draw [fill=red] (1.9,1.9) circle (2.5pt);
\draw [fill=red] (1.9,1.7) circle (2.5pt);
\draw [fill=red] (1.9,1.5) circle (2.5pt);
\draw [fill=red] (1.9,1.3) circle (2.5pt);
\draw [fill=red] (1.9,1.1) circle (2.5pt);
\draw [fill=red] (1.9,0.9) circle (2.5pt);
\draw [fill=red] (1.9,0.7) circle (2.5pt);
\draw [fill=red] (1.9,0.3) circle (2.5pt);
\draw [fill=red] (1.9,0.1) circle (2.5pt);
\draw [fill=red] (1.9,0.5) circle (2.5pt);
\draw [fill=red] (1.9,-1.9) circle (2.5pt);
\draw [fill=red] (1.9,-1.7) circle (2.5pt);
\draw [fill=red] (1.9,-1.5) circle (2.5pt);
\draw [fill=red] (1.9,-1.3) circle (2.5pt);
\draw [fill=red] (1.9,-1.1) circle (2.5pt);
\draw [fill=red] (1.9,-0.9) circle (2.5pt);
\draw [fill=red] (1.9,-0.7) circle (2.5pt);
\draw [fill=red] (1.9,-0.3) circle (2.5pt);
\draw [fill=red] (1.9,-0.1) circle (2.5pt);
\draw [fill=red] (1.9,-0.5) circle (2.5pt);
\end{tikzpicture}
        }}  &
        \subfloat[schematics for test case 2 domain] {\resizebox{0.4\columnwidth}{!}{
        \definecolor{red}{rgb}{1,0,0}
\definecolor{blue}{rgb}{0.08,0.4,0.75}
\begin{tikzpicture}[line cap=round,line join=round,>=triangle 45,x=1cm,y=1cm]
\clip(-3,-3) rectangle (5,3);

\draw [line width=0.8pt] (2,2)-- (2,-2);
\draw [line width=0.8pt] (2,-2)-- (-2,-2);
\draw [line width=0.8pt] (-2,-2)-- (-2,2);
\draw [line width=0.8pt] (-2,2)-- (2,2);

\draw [line width=0.4pt] (2,2) -- (2.7,2);
\draw [line width=0.4pt] (2,-2) -- (2.7,-2);
\draw [line width=0.4pt] (2,-2) -- (2,-2.7);
\draw [line width=0.4pt] (-2,-2) -- (-2,-2.7);

\draw [<->,line width=0.4pt] (2.5,2) -- (2.5,-2);
\draw [<->,line width=0.4pt] (2,-2.5) -- (-2,-2.5);

\draw (0,-2.5) node[anchor=north] {$L=1200$};
\draw (2.5,0) node[anchor=west] {$L=1200$};

\draw [fill=red] (-1.9,-1.9) circle (2.5pt);
\draw [fill=red] (-1.9,1.9) circle (2.5pt);
\draw [fill=red] (1.9,-1.9) circle (2.5pt);
\draw [fill=red] (1.9,1.9) circle (2.5pt);
\draw [fill=blue] (0,0) circle (2.5pt);

\end{tikzpicture}
        }}
    \end{tabular}
    \caption{the producers and injectors are highlighted with red and blue colors, respectively. parameters ($w, l_1 \textup{ and } l_2$) for test case 1 log permeability are shown in fig (a), where high permeability channel is colored in gray. }
    \label{fig:domain_schema}
\end{figure}

\textit{Test case 1 (Channel like permeability distribution)}: Figure \ref{fig:domain_schema}a shows schematic of the first test case (inspired from the case study done by \cite{brouwer2001recovery}). 
A total number of 31 injectors are placed on the left edge of the domain while an equal number of producers are placed symmetrically on the right side. 
A linear high permeability channel  connects from left to right side of the domain. 
The channel location is parameterized with its left and right distance ($l_1$ and $l_2$) from the top and channel width $w$. 
These parameters follow uniform distributions defined as,  $w \sim U(120, 360)$, $l_1 \sim U(0,L-w)$ and $l_2 \sim U(0,L-w)$, where $L$ is domain length. 
log permeability $g$ is sampled from the uncertainty distribution $\mathcal{G}_1$:
\begin{equation*}
    g \sim \mathcal{G}_1(w, l_1, l_2).
\end{equation*}
To be specific, log permeability $g$ at a location $(x,y)$ is formulated as:
\begin{equation*}
    g(x,y) = \left\{\begin{matrix}
g_{\textup{high}} & \textup{if} & \frac{l_2-l_1}{L}x+l_1 \leq y \leq \frac{l_2-l_1}{L}x+l_1+w, \\ 
 &  & \\ 
g_{\textup{low}} & \textup{otherwise}, & 
\end{matrix}\right.
\end{equation*}
where $x$ and $y$ are horizontal and vertical distances from the upper left corner of the domain illustrated in figure \ref{fig:domain_schema}a.
The values for $g_{high}$ and $g_{low}$ (5.5 and -2,  respectively) are inspired from \textit{Upperness} permeability distribution specified in SPE-10 model 2 case (refer to \ref{app: k_dist_params} for details).
Figure \ref{fig:eval_k}a illustrate various samples drawn from the distribution $\mathcal{G}_1$.

\textit{Test case 2 (Spatially correlated smooth permeability distribution)}: We use test case 2 to represent uncertainty distribution of a smoother permeability field. Figure \ref{fig:domain_schema}b illustrates reservoir domain for this case. 
It comprises of four producers located at four corners of the domain and an injector located at the center of the domain. 
The permeability distribution for this case is considered as a log normal distribution which is constrained with fixed values at well locations.
As a result, log permeability $g$ is sampled from the normal distribution $\mathcal{G}_2$:
\begin{align}
    g  & \sim \mathcal{G}_2(\mu_g, \Sigma_g), \textup{ where,} \label{eq: g2_dist} \\
    \mu_g &= 2.41, \label{eq: g2_mean} \\
    \Sigma_g &= C(x,x) - C(x,x'){C(x', x')}^{-1}C(x, x'), \label{eq: g2_sigma}
\end{align}
where, $C(x,x')$ is the co-variance matrix between unconstrained domain locations $x$ and constrained locations $x'$ while $\mu_g$ correspond to the constrained log-permeability value at the well locations. 
The co-variance matrix is calculated using an exponential kernel as:
\begin{equation}
    C(a,b) = \sigma^2 \exp{ - \frac{\left \| a -b  \right \|}{l}  }.
    \label{eq: case2_corr_len}
\end{equation}
We choose correlation length $l$ as 240 ft (20\% of domain length) and the variance amplitude $\sigma$ as 2.5. 
The values $\mu_g$, $\sigma$ and $l$ were chosen to fit permeability distribution in the same range of that in \textit{Tarbert} case specified in SPE-10 model 2 case (refer \ref{app: k_dist_params}). 
Examples of samples drawn from the distribution $\mathcal{G}_2$ are shown in figure \ref{fig:eval_k}b.

\subsection{RL problem formulation:}
Both, PPO and A2C, algorithms attempt to learn neural network parameters $\theta$ to learn the policy $\pi_{\theta}(a|s)$. 
We choose five step episode which is obtained by dividing the temporal domain $\mathcal{T}$ into five control steps. 
The optimisation potential of the problem can be improved if we choose a higher than 5 control steps. However, we choose 5 control steps for the ease of execution and demonstration.
The episode steps are denoted with $t_m$ where $m \in \{1,2,\cdots, 5 \}$. State is represented by observation  $o(x',t_m)$ which is a vector of saturation and pressure values at all well locations.
However, since the saturation at injectors is always constant (one), we don't include it in the observation.
As a result, the observation vector is of the size $2n_p + n_i$ (i.e. 93 for test case 1 and 9 for test case 2) which forms the input of the policy network $\pi_{\theta}(a|s)$.
Action $a(x',t_m)$ is represented with a vector the size of number of control wells $n_p + n_i$ (i.e. 62 for test case 1 and 5 for test case 2). 
In order to maintain constraint defined in Eq.~\eqref{eq: constr}, we represent the action vector with weights, $w_i$s, such that $0.001\leq w_i \leq 1$ ( i.e. action vector is written as, $a(x',t_m)=( w_1,\cdots,w_{n_p},w_{n_p+1},\cdots,w_{n_p+n_i} )$). 
Using these weights, flow through $i$th producer, $a^-(x'_i,\cdot)$, is computed as,
\begin{equation*}
    a^-(x'_i,\cdot) = -\frac{w_i}{\sum_{i=1}^{n_p}w_i}c.
\end{equation*}
Similarly, flow through $i$th injector, $a^+(x'_i,\cdot)$, is written as,
\begin{equation*}
    a^+(x'_i,\cdot) = \frac{w_{i+n_p}}{\sum_{i=1}^{n_i}w_{i+n_p}}c.
\end{equation*}
The reward function defined in Eq.~\eqref{eq: reward_func} is normalized by dividing it with total pore volume ($\phi \times lx \times ly $) in order to obtain the reward in the range [0,1]. 
The normalized reward represents \textit{recovery factor} or \textit{sweep efficiency} for oil movement problem in petroleum reservoir.

\begin{figure}
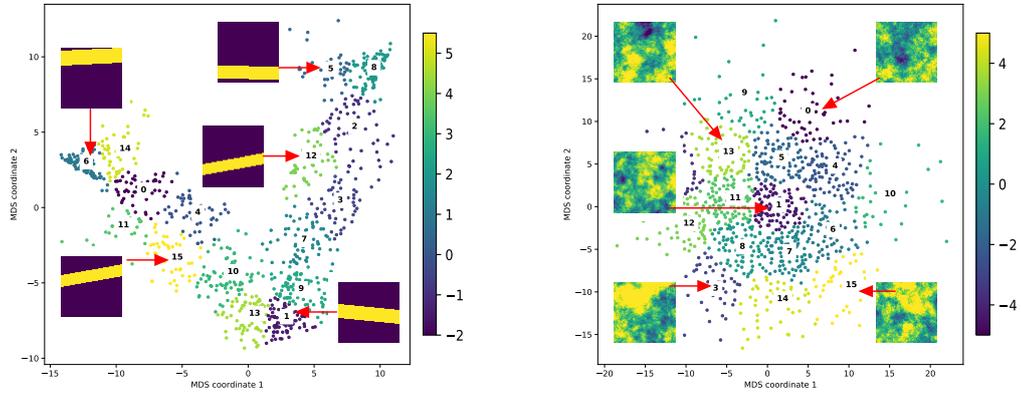

    \centering
    \begin{tabular}{cc}
        \subfloat[samples drawn from the $\mathcal{G}_1$ distribution] {\resizebox{0.45\columnwidth}{!} {
        \input{images/case_1_cluster_image.tikz}
        }}  &
        \subfloat[samples drawn from the $\mathcal{G}_2$ distribution] {\resizebox{0.45\columnwidth}{!}{
        \input{images/case_2_cluster_image.tikz}
        }}
    \end{tabular}
    \caption{clustering of log-permeability fields for test cases 1 and 2}
    \label{fig: cluster}
\end{figure}

\begin{figure}
    \centering
    \begin{tabular}{c c}
        \subfloat[samples of $\mathcal{G}_1$ in evaluation vector $\textbf{k}'$]{\includegraphics[width=0.45 \textwidth]{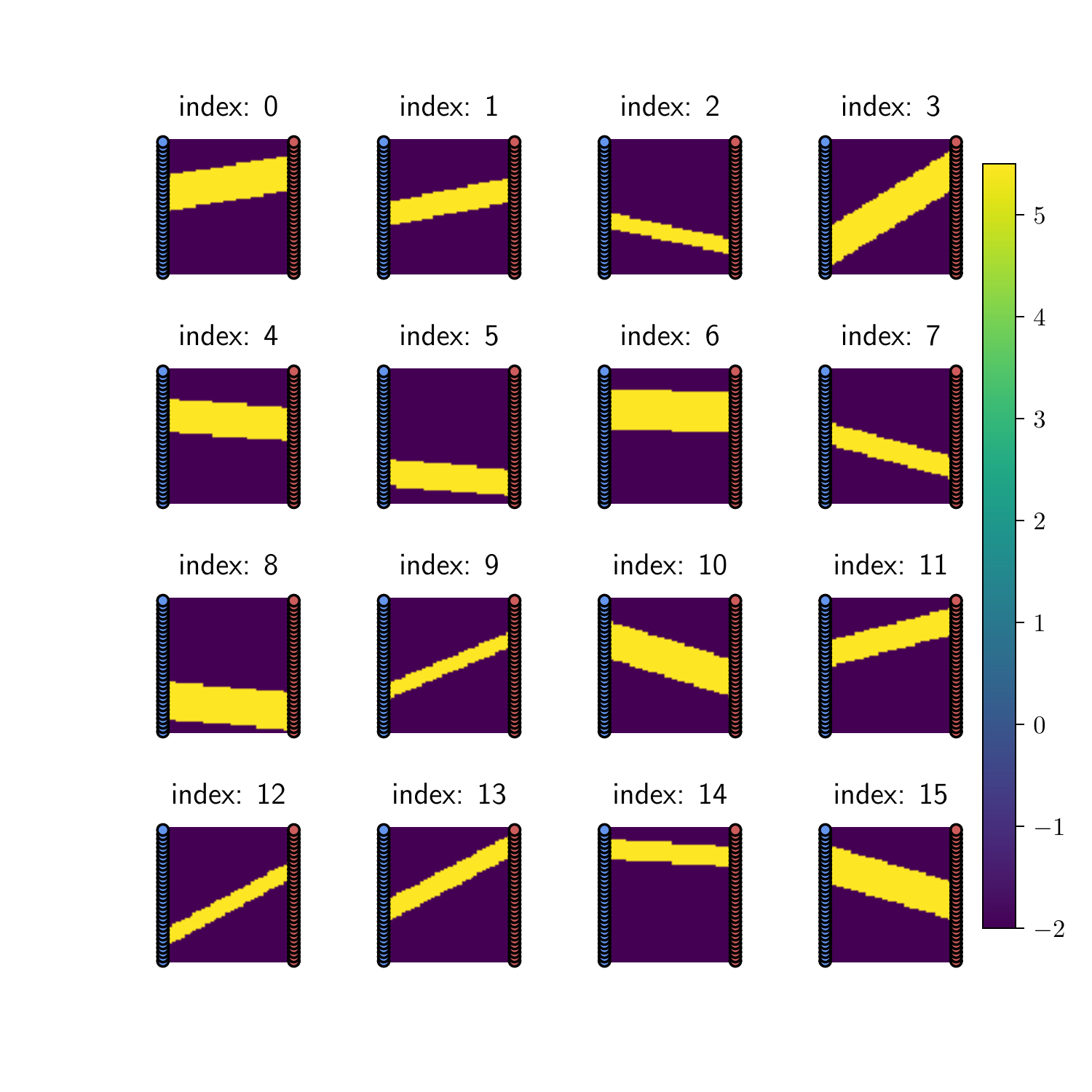}} 
        \subfloat[samples of $\mathcal{G}_2$ in evaluation vector $\textbf{k}'$]{\includegraphics[width=0.45 \textwidth]{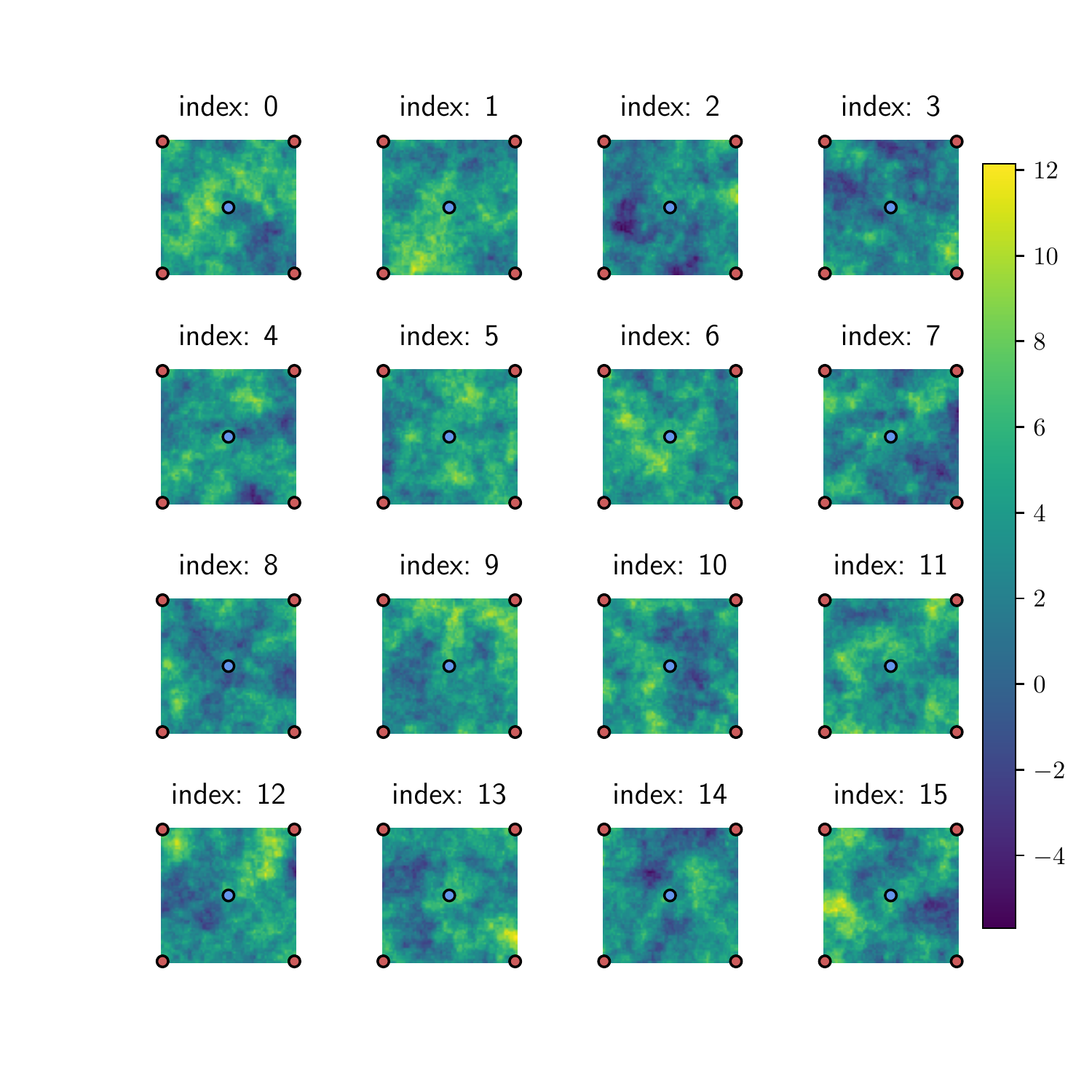}}
    \end{tabular}
    \caption{log-permeability plots for evaluation data for test cases 1 and 2}
    \label{fig:eval_k}
\end{figure}

We use clustering strategy explained in section \ref{sect: kmeans} where we choose total number of samples, $N$, and clusters, $l$, to be 1000 and 16 for both uncertainty distributions, $\mathcal{G}_1$ and $\mathcal{G}_2$. 
Training vector $\textbf{k}$ is obtained with samples $k_1, \cdots, k_{16}$ each corresponding to a cluster center.
Figure \ref{fig: cluster}a and \ref{fig: cluster}b show cluster plots for samples drawn from $\mathcal{G}_1$ and $\mathcal{G}_2$ permeability distributions, respectively. 
Permeability samples for test case 1 are distributed in the shape of an acute angle where samples in the vertex region correspond to high permeability channel at the central region, samples on the left arm correspond to high permeability channel in the upper region while samples in the right arm correspond to high permeability channel in the lower region of the domain. 
For test case 2, samples with more or less axisymmetric high permeability region are located in the central area in figure \ref{fig: cluster}b (e.g., cluster 1). 
The samples corresponding to eccentric high permeability regions are located outside as shown with examples $\textbf{k}_{13}$ (lower left region), $\textbf{k}_{3}$ (upper left region), $\textbf{k}_{15}$ (upper right region) and $\textbf{k}_{0}$ (lower right region) in figure \ref{fig: cluster}b. 
In order to represent well spread domain of $\mathcal{G}_1$ and $\mathcal{G}_2$ distributions, the 16 samples, each randomly chosen from a cluster, forms the evaluation vector $\textbf{k}'$. 
These evaluation samples are shown in figure \ref{fig:eval_k}a and \ref{fig:eval_k}b for test cases 1 and 2, respectively.

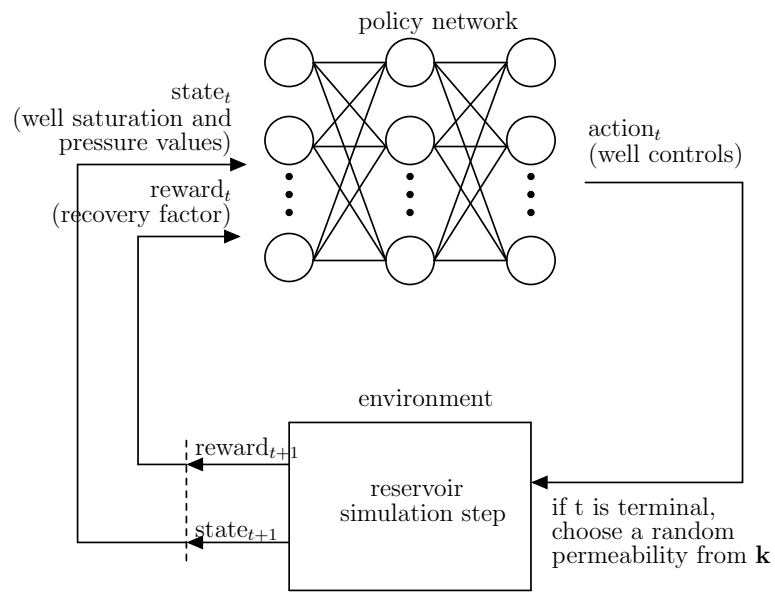
\begin{figure}
    \centering
    \begin{tabular}{c}
        \subfloat {\resizebox{0.7\columnwidth}{!} {
        \begin{tikzpicture}[line cap=round,line join=round,>=triangle 45,x=1cm,y=1cm]
\clip(-7,-6) rectangle (6,6);

\draw [line width=0.8pt] (2.9,2) -- (5.5,2);
\draw [line width=0.8pt] (5.5,2) -- (5.5,-3);
\draw [->,line width=0.8pt] (5.5,-3) -- (2,-3);
\draw [line width=0.8pt] (-4.5,-2.7) -- (-4.5,1.1);
\draw [->,line width=0.8pt] (-4.5,1.1) -- (-2.8,1.1);
\draw [line width=0.8pt] (-5.5,-4) -- (-5.5,2.3);
\draw [->,line width=0.8pt] (-5.5,2.3) -- (-2.7,2.3);
\draw [->,line width=0.8pt] (-2,-2.7) -- (-3.7,-2.7);
\draw [->,line width=0.8pt] (-2,-4) -- (-3.7,-4);
\draw [line width=0.8pt] (-4.5,-2.7) -- (-3.7,-2.7);
\draw [line width=0.8pt] (-5.5,-4) -- (-3.7,-4);
\draw [dashed, line width=0.8pt] (-3.7,-2.3) -- (-3.7, -4.4);

\draw [line width=0.8pt] (-2,-2)-- (-2,-4.8);
\draw [line width=0.8pt] (-2,-4.8)-- (2,-4.8);
\draw [line width=0.8pt] (2,-4.8)-- (2,-2);
\draw [line width=0.8pt] (2,-2)-- (-2,-2);

\draw [line width=0.8pt] (-1.6,4)-- (-0.4,2.6);
\draw [line width=0.8pt] (-1.6,4)-- (-0.4,4);
\draw [line width=0.8pt] (-1.6,4)-- (-0.4,0.7);
\draw [line width=0.8pt] (-1.6,2.6)-- (-0.4,4);
\draw [line width=0.8pt] (-1.6,2.6)-- (-0.4,2.6);
\draw [line width=0.8pt] (-1.6,2.6)-- (-0.4,0.7);
\draw [line width=0.8pt] (-1.6,0.7)-- (-0.4,4);
\draw [line width=0.8pt] (-1.6,0.7)-- (-0.4,2.6);
\draw [line width=0.8pt] (-1.6,0.7)-- (-0.4,0.7);
\draw [line width=0.8pt] (0.4,4)-- (1.6,4);
\draw [line width=0.8pt] (0.4,4)-- (1.6,2.6);
\draw [line width=0.8pt] (0.4,4)-- (1.6,0.7);
\draw [line width=0.8pt] (0.4,2.6)-- (1.6,4);
\draw [line width=0.8pt] (0.4,2.6)-- (1.6,2.6);
\draw [line width=0.8pt] (0.4,2.6)-- (1.6,0.7);
\draw [line width=0.8pt] (0.4,0.7)-- (1.6,4);
\draw [line width=0.8pt] (0.4,0.7)-- (1.6,2.6);
\draw [line width=0.8pt] (0.4,0.7)-- (1.6,0.7);
\draw [fill=black] (-2,1.5) circle (1.5pt);
\draw [fill=black] (-2,1.8) circle (1.5pt);
\draw [fill=black] (-2,2.1) circle (1.5pt);
\draw [fill=black] (0,1.5) circle (1.5pt);
\draw [fill=black] (0,1.8) circle (1.5pt);
\draw [fill=black] (0,2.1) circle (1.5pt);
\draw [fill=black] (2,1.5) circle (1.5pt);
\draw [fill=black] (2,1.8) circle (1.5pt);
\draw [fill=black] (2,2.1) circle (1.5pt);
\draw [line width=0.8pt] (-2,4) circle (0.4cm);
\draw [line width=0.8pt] (-2,2.7) circle (0.4cm);
\draw [line width=0.8pt] (-2,0.75) circle (0.4cm);
\draw [line width=0.8pt] (0,4) circle (0.4cm);
\draw [line width=0.8pt] (0,2.7) circle (0.4cm);
\draw [line width=0.8pt] (0,0.7) circle (0.4cm);
\draw [line width=0.8pt] (2,2.7) circle (0.4cm);
\draw [line width=0.8pt] (2,0.7) circle (0.4cm);
\draw [line width=0.8pt] (2,4) circle (0.4cm);

\draw (-0.8,-2.8) node[anchor=north west] {$\textup{reservoir}$};
\draw (-1.3,-3.2) node[anchor=north west] {$\textup{simulation step}$};
\draw (-1,5.0) node[anchor=north west] {$\textup{policy network}$};
\draw (-1,-1.3) node[anchor=north west] {$\textup{environment}$};
\draw (2.8,3.2) node[anchor=north west] {$\textup{action}_t$};
\draw (2.8,2.8) node[anchor=north west] {$\textup{(well controls)}$};
\draw (-3.7,-2.1) node[anchor=north west] {$\textup{reward}_{t+1}$};
\draw (-3.7,-3.5) node[anchor=north west] {$\textup{state}_{t+1}$};
\draw (2.2,-3.1) node[anchor=north west] {$\textup{if t is terminal,}$};
\draw (2.2,-3.5) node[anchor=north west]
{$\textup{choose a random}$};
\draw (2.2,-3.9) node[anchor=north west]
{$\textup{permeability from \textbf{k}}$};

\draw (-2.8,2.2) node[anchor=north east] {$\textup{reward}_t$};
\draw (-2.8,1.8) node[anchor=north east] {$\textup{(recovery factor)}$};

\draw (-2.8,3.8) node[anchor=north east] {$\textup{state}_t$};
\draw (-2.8,3.4) node[anchor=north east] {$\textup{(well saturation and}$};
\draw (-2.8,3) node[anchor=north east] {$\textup{pressure values)}$};


\end{tikzpicture}
        }} 
    \end{tabular}
    \caption{RL algorithm agent-environment interaction schematics to learn robust optimal well control policy}
    \label{fig: rl_schema}
\end{figure}

Figure \ref{fig: rl_schema} outlines the general schematics for agent-environment interactions in PPO and A2C algorithms for robust optimal control problem. 
Since these algorithms are stochastic in nature, we provide training and evaluation returns as a mean corresponding to three distinct seed values.
These results are benchmarked against DE algorithm optimisation results.
Parameters used for all algorithms along with the confidence range of learning plots are presented in \ref{app: rl_params}.

\section{Results and discussion} \label{sect: results}
\begin{figure}
    \centering
    \includegraphics[width=\textwidth]{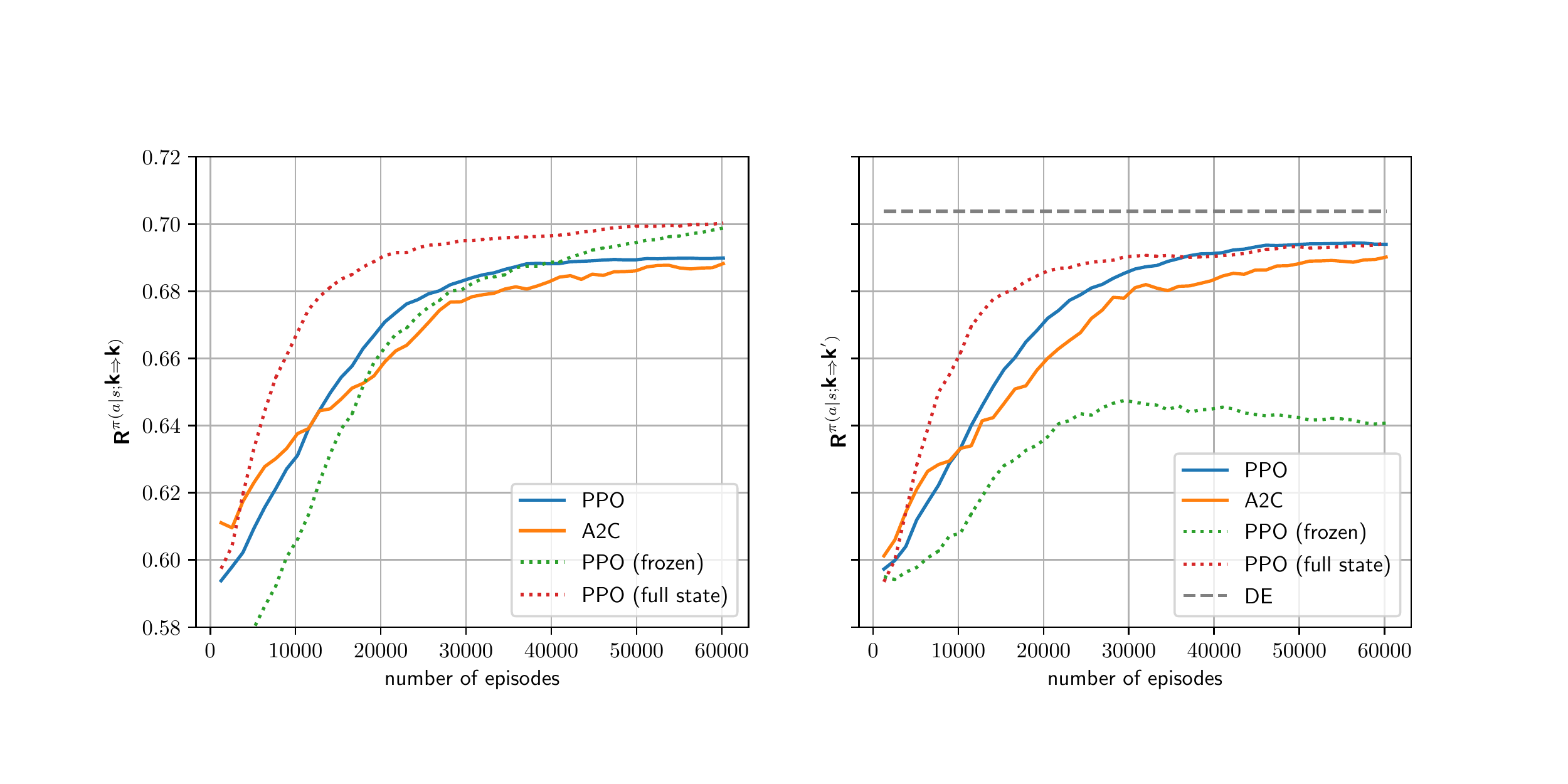}
    \caption{Test case 1: monitoring plots for average training return $\textbf{R}^{\pi(a|s; \textbf{k} \Rightarrow \textbf{k})}$ (on left) and evaluation return $\textbf{R}^{\pi(a|s; \textbf{k} \Rightarrow \textbf{k}')}$ (on right) for learning process in PPO, A2C and frozen PPO. The evaluation return value is compared with the optimisation results obtained using DE  }
    \label{fig: learning_Case1}
\end{figure}

We refer to the control policy in which all wells are equally open as the \textit{base} policy. 
When the first test case is operated with base policy, most of the water flooding take place in the high permeability channel causing poor sweep efficiency in the low permeability region. 
Naturally, the optimal policy is to restrict the flow through wells which are in the region nearby high permeability channel. 
Using DE algorithm, we obtain 16 optimized solutions for optimal control problem defined in Equations \eqref{eq: prob_def} where each value in evaluation vector $\textbf{k}'$ is treated as fixed permeability $k$. These results act as a reference to optimal solutions obtained using PPO and A2C algorithms. 
Reference value for mean evaluation return $\textbf{R}^{\pi^*(a|s; \textbf{k} \Rightarrow \textbf{k}')}$ is obtained by averaging DE results of these 16 problems.
Note that DE algorithm is not a suitable method to solve the robust optimal control problem since it can provide optimal controls only for certain permeability samples as opposed to PPO or A2C algorithms where we try to learn the policy that is applicable to all samples of permeability distribution.
However, DE results are used as the reference since they provide the upper bounds achieved by direct optimization on sample by sample basis.
Figure \ref{fig: learning_Case1} shows plots for training ($\textbf{R}^{\pi(a|s; \textbf{k} \Rightarrow \textbf{k})}$) and evaluation ($\textbf{R}^{\pi(a|s; \textbf{k} \Rightarrow \textbf{k}')}$) returns versus total number of episodes for PPO and A2C learning process. 
As can be seen, PPO and A2C algorithms successfully learned the robust optimal policy and their average evaluation returns $\textbf{R}^{\pi^*(a|s; \textbf{k} \Rightarrow \textbf{k}')}$ are within the range of DE results. 
We also present results for a frozen PPO policy trained using a fixed permeability located at index 1 in the training vector \textbf{k} (indicated with dotted green line in figure \ref{fig: learning_Case1}). 
We note that the training vector \textbf{k} for frozen PPO case only comprises of a single permeability realization.
This frozen policy is not robust as it performs poorly on unseen permeabilities as demonstrated when we plot $\textbf{R}^{\pi(a|s; \textbf{k} \Rightarrow \textbf{k}')}$ value in its learning process. 
Furthermore, learning plot for PPO algorithm with full state representation is illustrated with red dotted line.
In this case, we provide the agent with saturation values at each grid point in the domain (i.e. with a vector of length $61\times61=3721$).
Policy learning with full state information are in the same range of that with only well observation state representation.
In other words, information of well observations is enough to form optimal policy for this case.

\begin{figure}
    \centering
    \includegraphics[width=\textwidth]{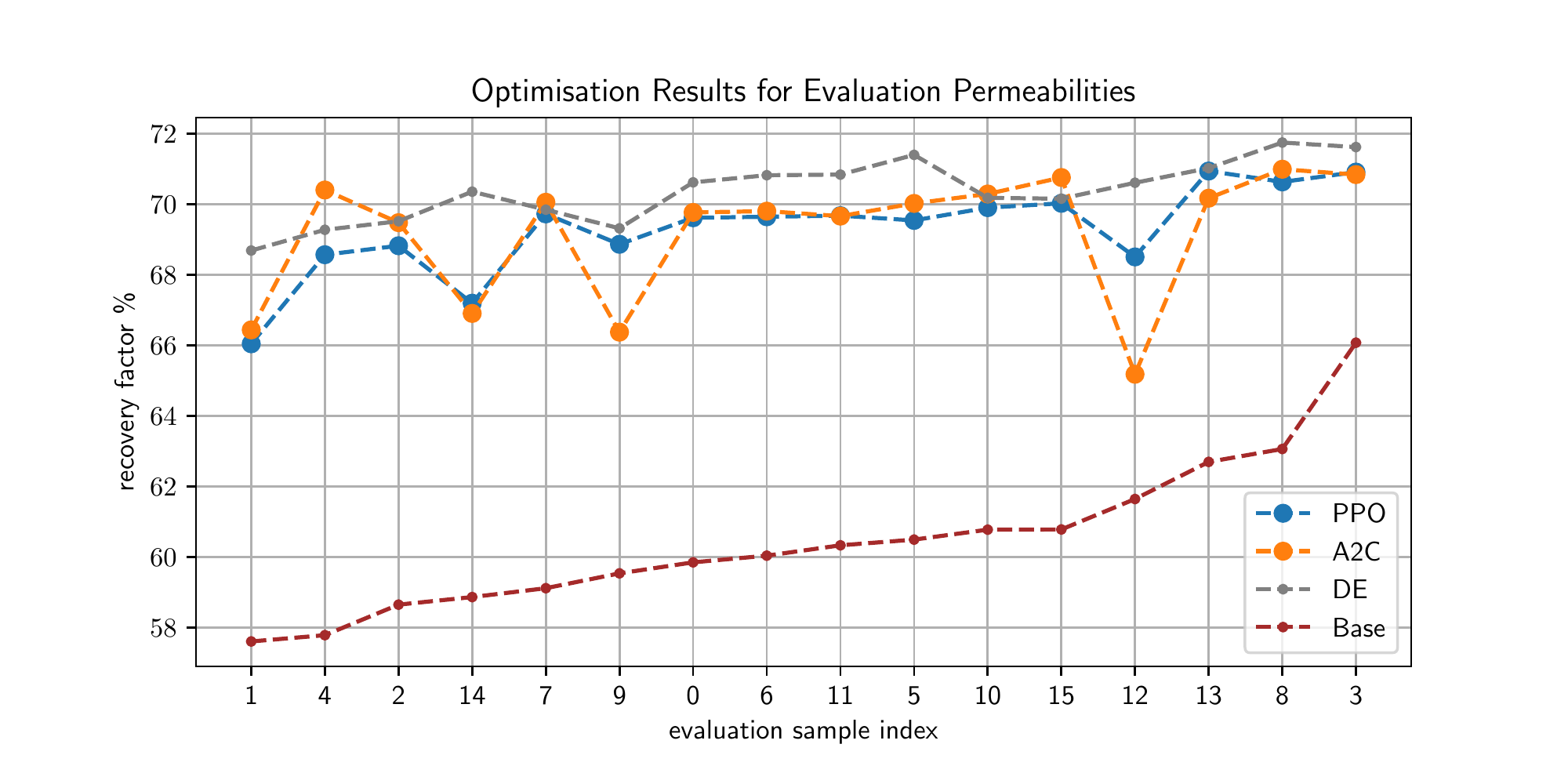}
    \caption{Test case 1: comparison of optimum recovery factor (in $\%$) for each permeability value from evaluation vector $\textbf{k}'$. Results of of PPO, A2C and DE are compares with base control actions (all control valves equally open) results }
    \label{fig:eval_res_case_1}
\end{figure}

\begin{figure}
    \centering
    \begin{tabular}{c}
        \subfloat[Evaluation permeability index 3]{\includegraphics[width=0.6\textwidth]{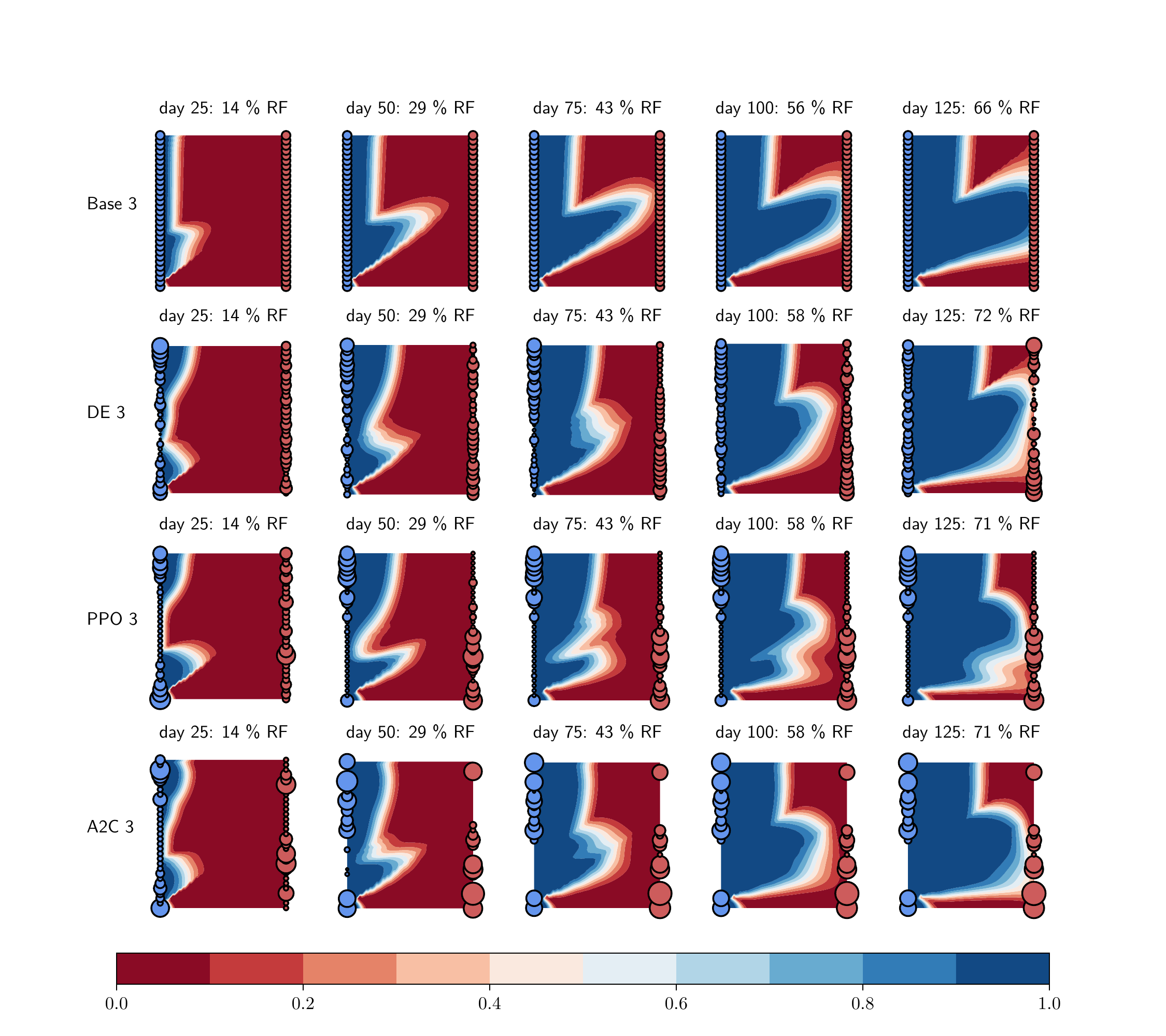}} \\
        \subfloat[Evaluation permeability index 5]{\includegraphics[width=0.6\textwidth]{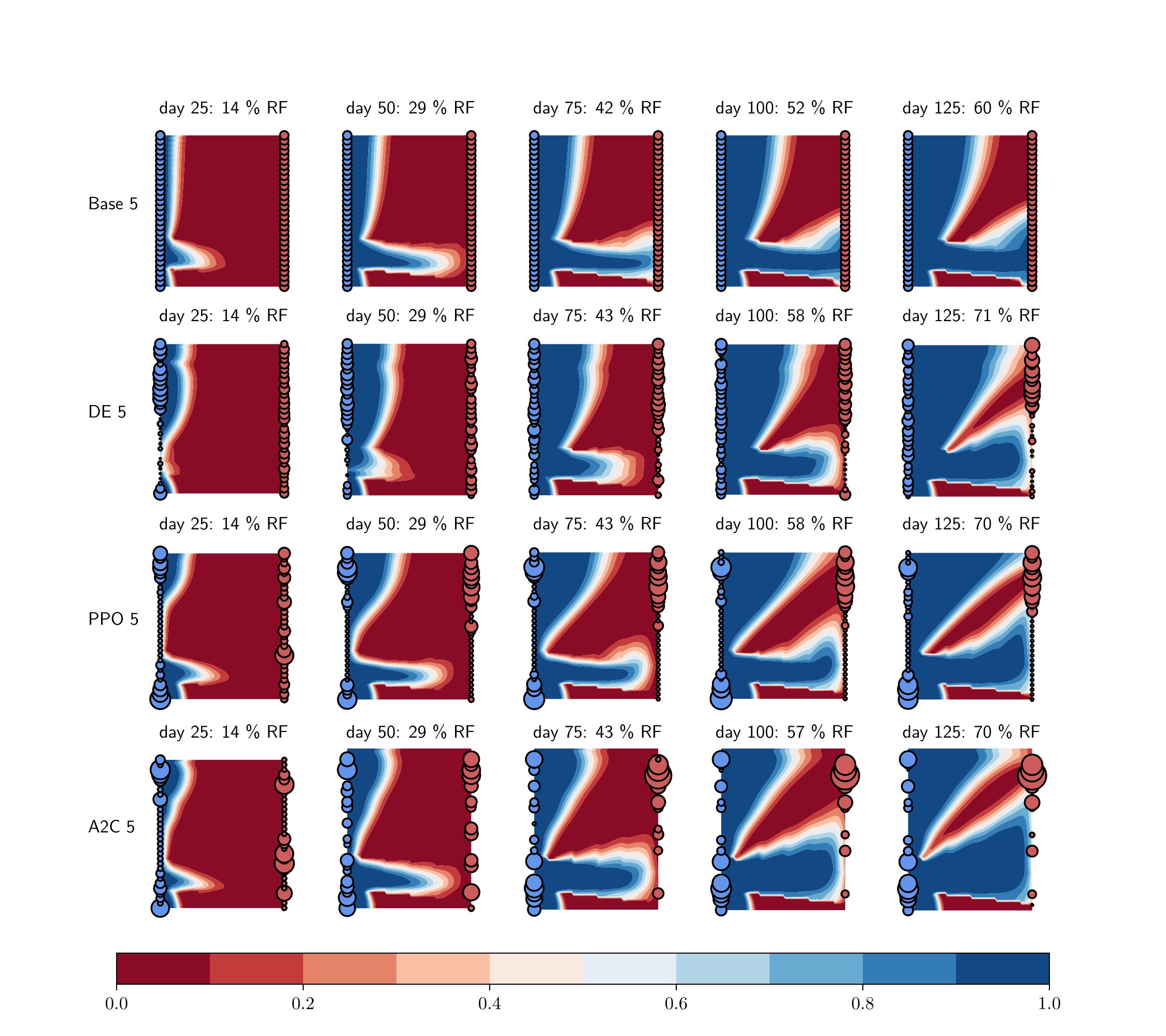}} \\
    \end{tabular}
    \caption{Test case 1: optimal well controls for permeability values $\textbf{k}_3'$and $\textbf{k}'_5$. Producer and injector flow controls are indicated with red and blue circles, respectively. Values of flow controls are proportional to the circle radius. }
    \label{fig:ws_case1}
\end{figure}

Figure \ref{fig:eval_res_case_1} plots optimum recovery factors (in $\%$) corresponding to each evaluation permeability in the vector $\textbf{k}'$.
Results of PPO and A2C policy are comparable to the DE results which are independently optimized for each permeability field. 
Figure \ref{fig:ws_case1} illustrates the optimum controls corresponding to evaluation permeability at third and fifth indices of $\textbf{k}'$. 
The controls for injectors and producers are shown in blue and red colored circles, respectively. 
Note that the radius of the circle at certain well location is proportional to flow through that well.
That is, the radius of the circle at certain well location is proportional to the flow control opening of the corresponding well. 
As can be seen in figure \ref{fig:ws_case1}, the optimal control policy to restrict flow controls in the high permeability region is successfully learned using PPO and A2C.

In the second test case, all reservoir parameters and well locations are axisymmetric except the permeability field. 
As a result, there is an imbalance in the flow direction from the central injector. 
The optimal flow control policy is to govern the well controls so that balanced amount of sweeping can be maintained in all four quadrants of the reservoir. 
For instance, if water sweeps uniformly in all quadrants of reservoir except the upper left, the optimal policy should increase the flow through upper left producer while restricting the flow through rest of the producers (i.e. govern the controls to cope with the imbalance in the upper left quadrant). 
For the five-spot case under consideration, the optimal policy has in total 10 modes: four due to imbalance in single quadrant and six due to imbalance in a pair of quadrants.

\textit{confusion in RL policy learning}: Policy $\pi_{\theta}(a|s)$ is learned through numerous agent-environment interactions experienced with 16 permeability field instances in training vector \textbf{k}. 
By definition, the optimal policy returns the action $a$ that corresponds to maximum return episode from current state $s$. 
Since the first state $s_0$ is same for all permeability fields (according to initial condition defined in Eq.~\eqref{eq: init_eq}), the first action $\pi_{\theta}(a_0|s_0)$ is always the one that correspond to a certain permeability field in training vector \textbf{k} which produces maximum total return.
So if we imagine $\textbf{k}_{5}$ to be such a permeability which happens to follow one of the ten optimal policy modes (say mode 7). 
The first action $a_1$ will always be the one that correspond to mode 7 policy. 
This is obviously undesirable when we apply this policy on permeability fields which correspond to another mode of optimal policy. 
In order to avoid this \textit{confusion in policy learning}, we train RL policies from second step onward. 
Since the second step of the episode is different for different permeability fields, RL policy learning doesn't face this confusion anymore. 
By default we treat the first action to follow base policy (i.e. all wells open equally). 
Note that the first action for test case 1 RL optimal policies is also identical for all cases (refer to figure \ref{fig:ws_case1}). 
However, since the optimal policy's nature is not modal, the resulting sub-optimality is not as much prominent for this case.

\begin{figure}
    \centering
    \includegraphics[width=\textwidth]{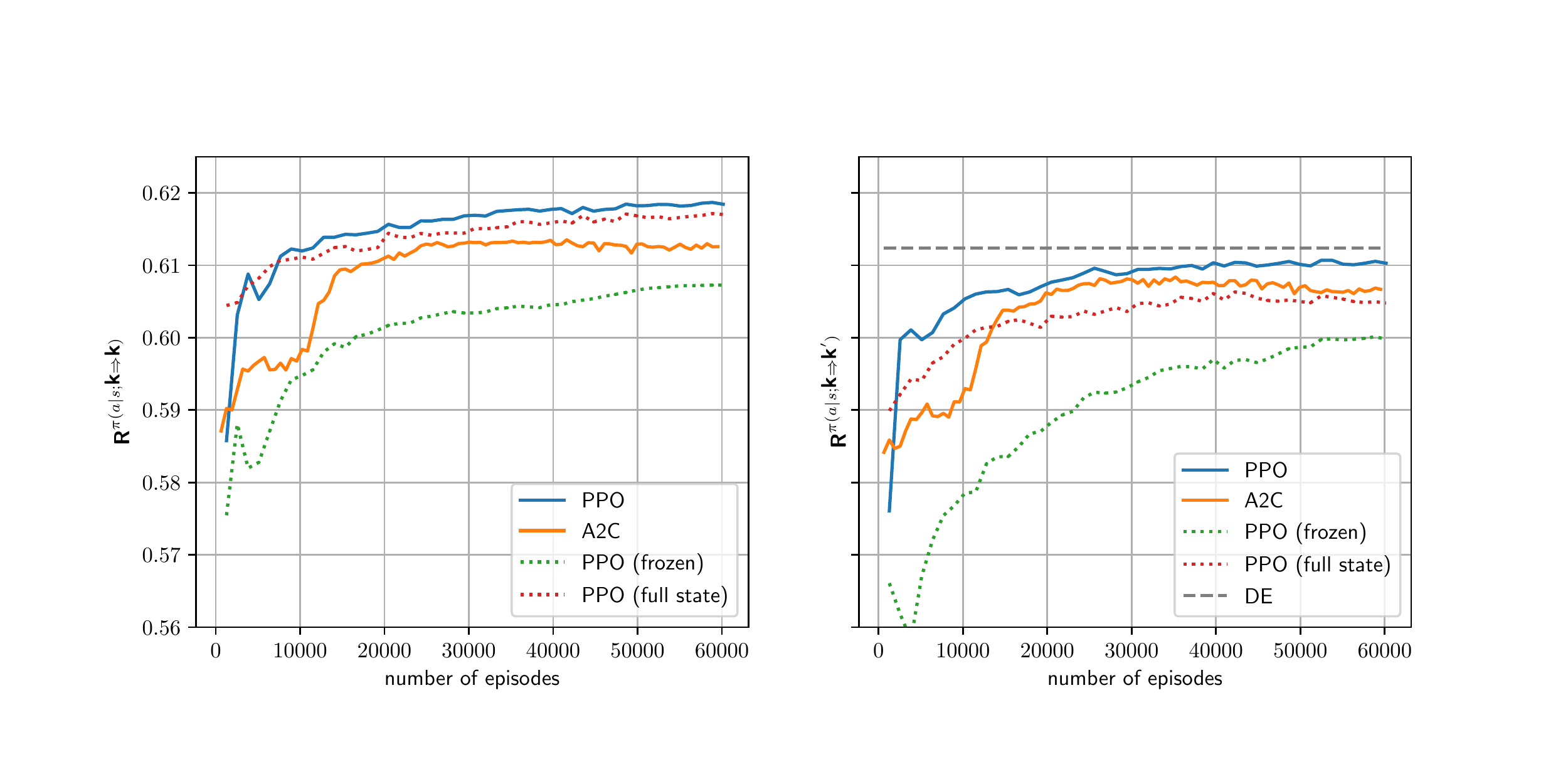}
    \caption{Test case 2: monitoring plots for average training return $\textbf{R}^{\pi(a|s; \textbf{k} \Rightarrow \textbf{k})}$ (on left) and evaluation return $\textbf{R}^{\pi(a|s; \textbf{k} \Rightarrow \textbf{k}')}$ (on right) for learning process in PPO and A2C. The evaluation return value is compared with the optimisation results obtained using DE  }
    \label{fig:learning_Case2}
\end{figure}

\begin{figure}
    \centering
    \includegraphics[width=\textwidth]{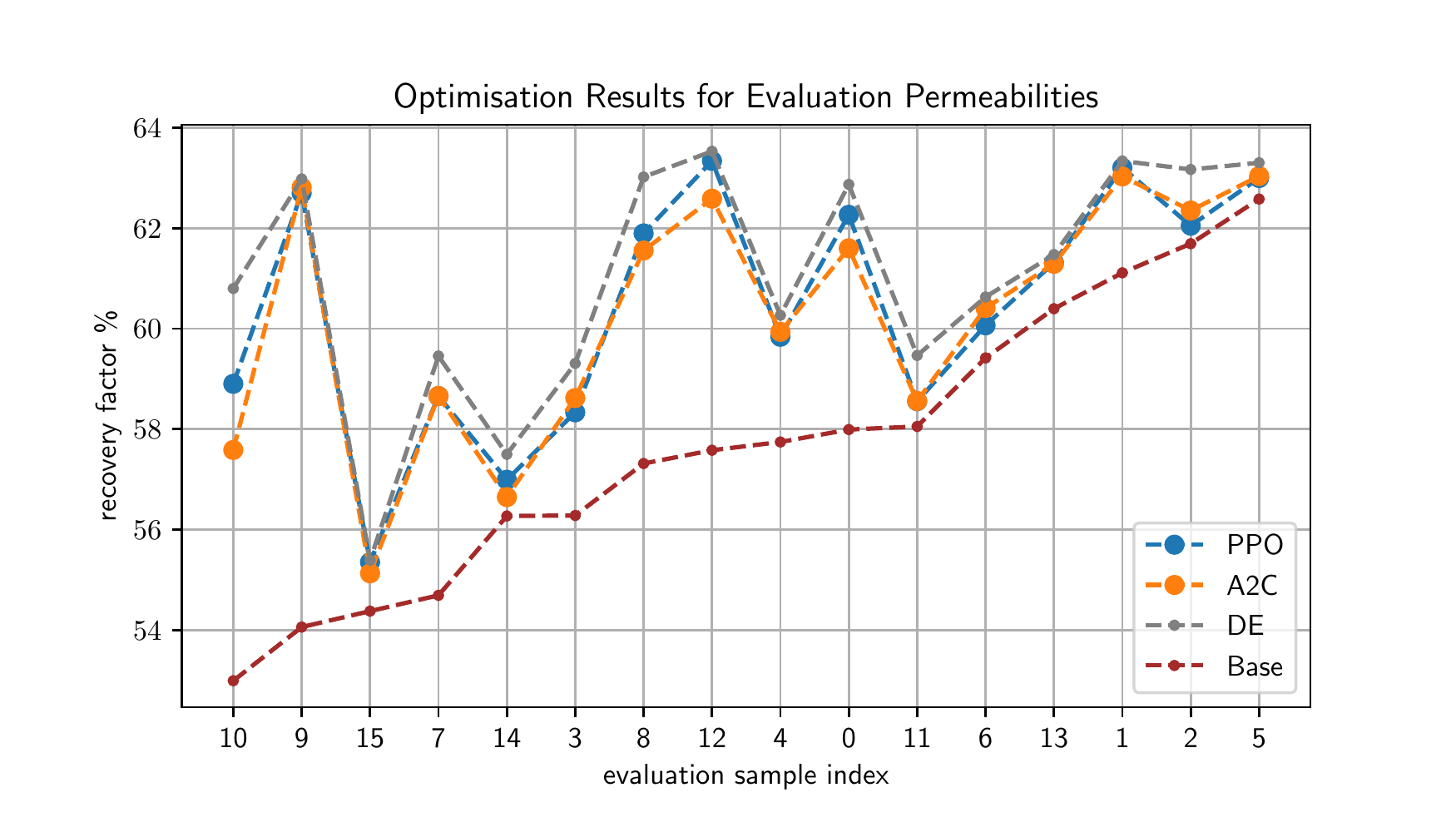}
    \caption{Test case 2: comparison of optimum recovery factor (in $\%$) for each permeability value from evaluation vector $\textbf{k}'$. Results of of PPO, A2C and DE are compared with base control actions (all control valves equally open) results }
    \label{fig:eval_res_case_2}
\end{figure}
\begin{figure}
    \centering
    \begin{tabular}{c}
        \subfloat[Evaluation permeability index 9]{\includegraphics[width=0.6\textwidth]{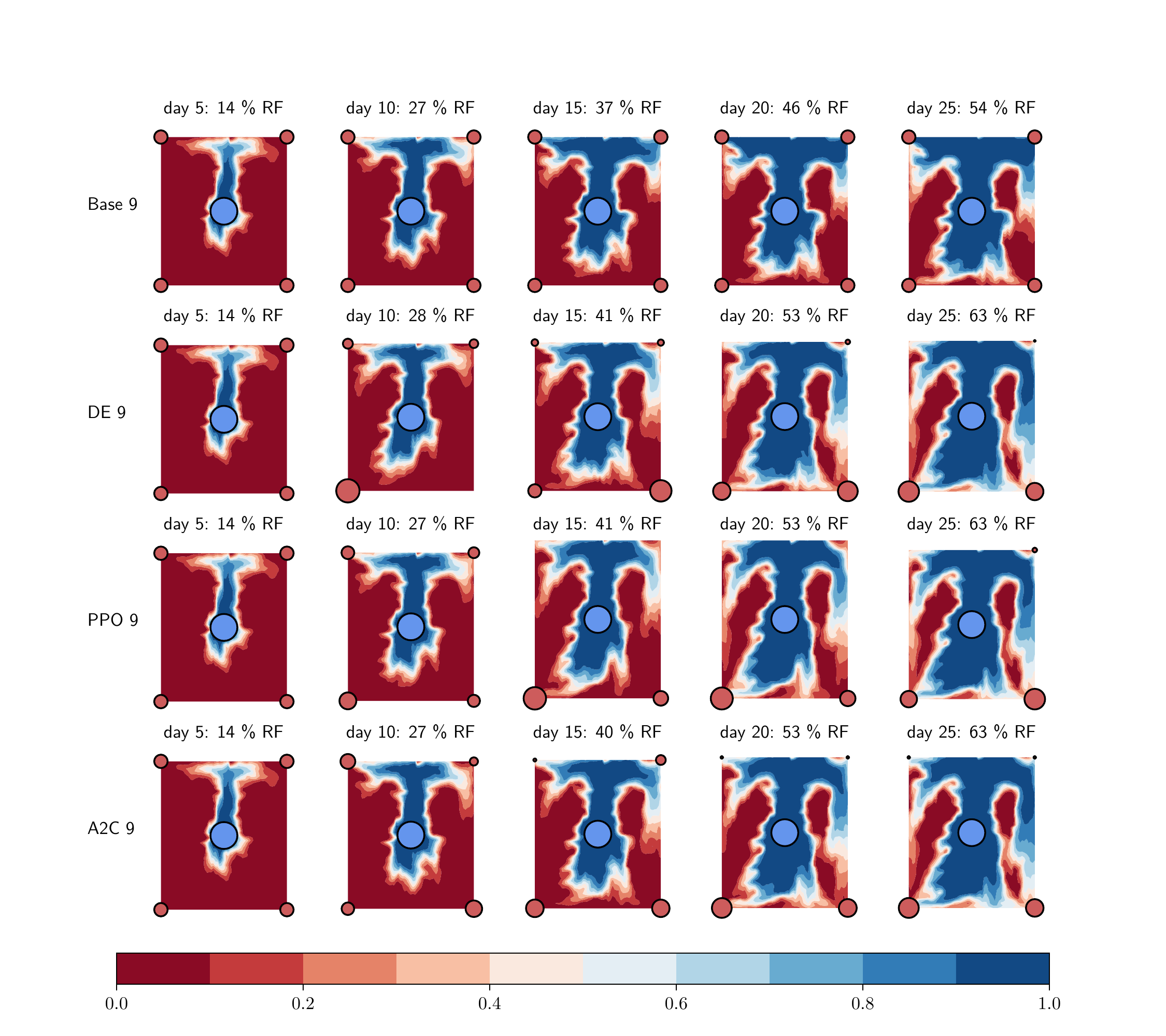}} \\
        \subfloat[Evaluation permeability index 12]{\includegraphics[width=0.6\textwidth]{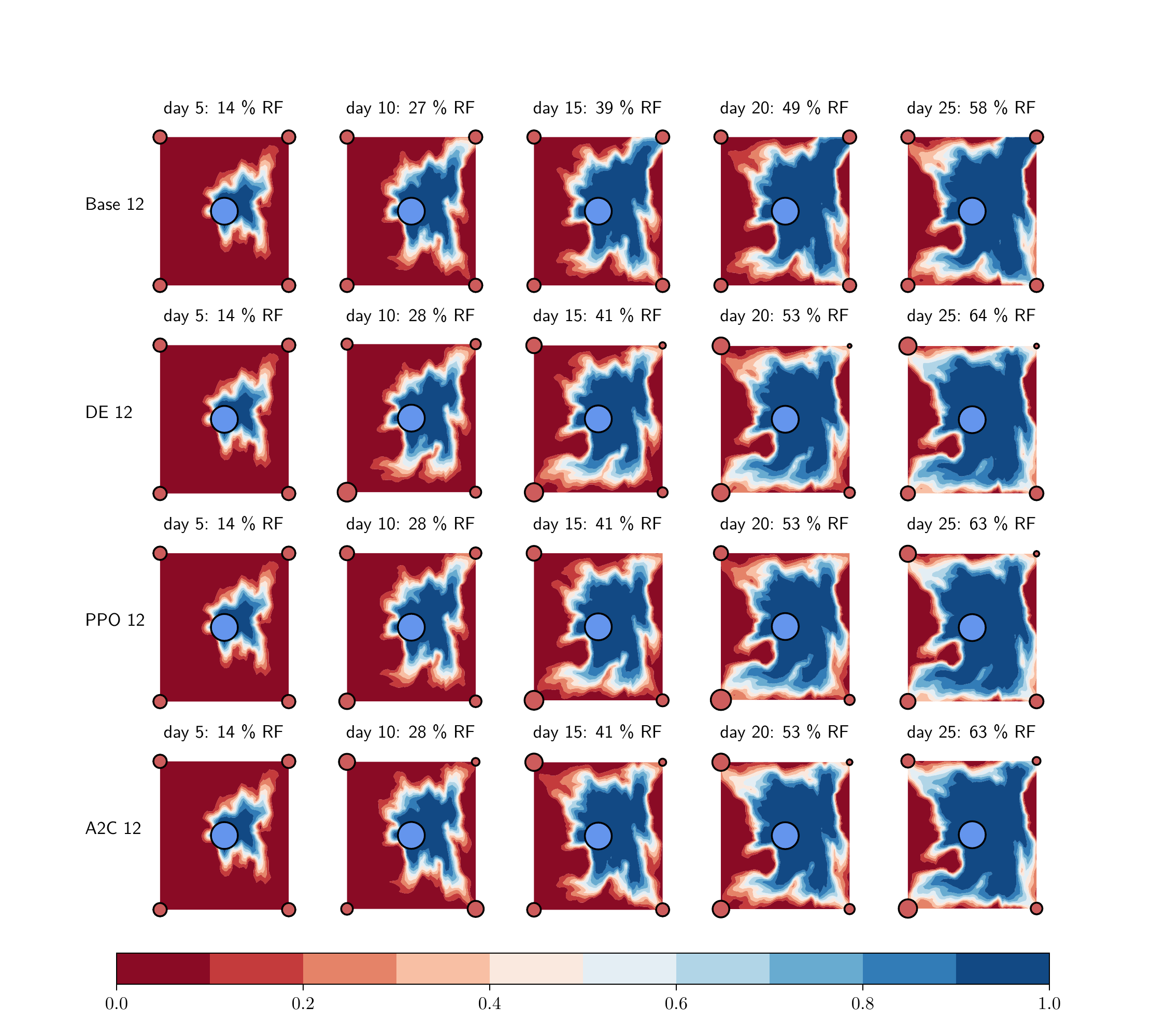}} \\
    \end{tabular}
    \caption{Test case 2: optimal well controls for permeability values $\textbf{k}'_9$ and $\textbf{k}'_{12}$. Producer and injector flow controls are indicated with red and blue circles, respectively. Values of flow controls are proportional to the circle radius.}
    \label{fig:ws_case2}
\end{figure}

Figure \ref{fig:learning_Case2} shows learning plots for training and evaluation returns for PPO, A2C and DE algorithms.
Similar to test case 1 results, PPO and A2C algorithms successfully learn robust optimal policy.
RL policies learned with well observations show results in the same range with the PPO policy with full state representation.
As expected, the frozen policy's lack of robustness can be seen in evaluation return learning plot.
PPO, A2C and DE optimisation results for evaluation permeability fields in $\textbf{k}'$ are compared individually in figure \ref{fig:eval_res_case_2}.
RL policies successfully capture the optimal policy behaviour as experienced with DE results. 
Figure \ref{fig:ws_case2} illustrate control policies for evaluation permeability fields at ninth and twelfth indices of $\textbf{k}'$. 
For instance, optimal policy for $\textbf{k}'_9$ which refers to increasing flow through producers in the lower region of the domain is clearly observed in its RL policies.

Computational complexity is the major limitation of using RL in simulation based control problem. For all three optimisation methods (PPO, A2C and DE), we employ multiprocessing to reduce total computational time. 
RL algorithms use multiple CPUs to run episodes in parallel while GPUs are used for neural network back propagation computations. 
Depending on resource availability and parameter tuning, different number of CPUs and GPUs are used on case basis (refer \ref{app: rl_params} for number of CPU used in each case). Furthermore, parallelism behaviour is also varied between RL and DE algorithms. 
In RL algorithms, simulations are run in parallel just as in the case of DE. However, in RL algorithms, neural networks are backpropagated synchronously at the end of each iteration which causes extra computational time in waiting and data distribution. 
In order to compare computational efforts irrespective of computational resources and parallelism behaviours, it is therefore, fair to compare number of simulation runs which is a major source of computational cost. 
For both cases, RL algorithms were let run for 60000 episodes which correspond to 60000 simulation runs and each such algorithm was run for three seed values (In total $3\times60000=180000$ simulation runs). 
For the first test case, DE algorithms consisted of 750 generations each comprising 305 population size and since DE algorithm was used for all 9 samples in $\textbf{k}'$, the total number of simulation runs is 2058750 ($750 \times 305 \times 9$). 
Similarly, for the second test case, DE comprised of total 135000 simulation runs ($750 \textup{ generations} \times 20 \textup{ population size} \times 9 \textup{  samples}$). 
Parameters and computational resources used for all algorithms are delineated in \ref{app: rl_params}. 

\begin{table}
    \caption{number of simulation runs in each algorithm}
    \centering
    \begin{tabular}{l l l}
        \hline
         & Test case 1 & Test case 2  \\
        \hline
        PPO & 180000 & 180000\\
        A2C & 180000 & 180000\\
        DE & 2058750 & 135000\\
        \hline
    \end{tabular}
    \label{tab:sim_runs}
\end{table}

\section{Conclusions} \label{sect: conclusion}
We present a case study for using model-free RL algorithms to obtain robust optimal control policy for optimal well control problems.
This policy is learned under the assumption that the system is partially observable and is governed by a system of nonlinear partial differential equations. 
The robustness of these policies were obtained using a domain randomisation scheme that uses only few samples from a predefined uncertainty distribution by utilizing cluster analysis.
Further, the optimality of these policies were successfully benchmarked against reference solutions obtained by direct optimisation using DE algorithm. 
We consider the current framework as a first attempt towards application of narrow AI to the field of subsurface flow control where data is only available at the well locations. 

In the current study we made the following key assumptions:
\begin{itemize}
    \item the optimal control problem is formulated in the form of Eq. \eqref{eq: prob_def} which comprises of an objective function (Eq. \eqref{eq: obj_fun}), a governing equation (Eq. \eqref{eq: gov_eq}), initial/boundary conditions (Eq. \eqref{eq: init_eq}) and constraints (Eq. \eqref{eq: constr}),
    \item parameter uncertainty distribution is predefined,
    \item transition between the partial observations of the system approximately follows the Markov property,
    \item effectiveness of the optimal policy $\pi^*(a|s;k\Rightarrow k')$, is inversely related to the distance $D(k, k')$.
\end{itemize}
As a result similar techniques to those presented here could be applied to other simulation-based applications as long as these assumptions are met.

In the current study, we train RL policies with a large number of simulation runs. This can be computationally intractable for large scale models with long simulations run times. In future studies, we aim to address this issue by utilizing fast surrogate modeling techniques to accelerate the reinforcement learning process.

\section*{Acknowledgement}
The first author would like to acknowledge the Ali Danesh scholarship to fund his PhD studies at Heriot-Watt University. The authors would like to acknowledge the EPSRC funding through the grant EP/V048899/1.

\clearpage
\appendix

\section{Definition of value and advantage functions} \label{app: value_funs}
In RL, the policy $\pi(a|s)$ is said to be optimal if it maps the state $s_t$ with an action $a_t$ that correspond to maximum expected return value. These return values are learned through the data obtained in agent-environment interactions. Following are some definition of return values typically used in RL:

\textit{Value function} is the expected future return for a particular state $s_t$ and is defined as,
\begin{equation*}
    V(s) = \mathbb{E}_{\pi} \left [ \sum_{m} \gamma^m r_{m+t+1} | s_t = s \right],
\end{equation*}
where $\mathbb{E}_{\pi}[\cdots]$ denotes expected value given that the agent follows the policy $\pi$. As short hand, we denote $V(s)$ at state $s_t$ as $V_t$.

\textit{Q function} is similar to value function except that it represent the expected return when the agent takes action $a_t$ in the state $s_t$. It is defined as,
\begin{equation*}
    Q(s, a) = \mathbb{E}_{\pi} \left [ \sum_{m} \gamma^m r_{m+t+1} | s_t = s, a_t = a \right].
\end{equation*}
\textit{Advantage function} is defined as the difference between Q function and value function and is denoted by $A_t$ at state $s_t$ and action $a_t$.

\section{Permeability uncertainty distribution parameters} \label{app: k_dist_params}
\begin{figure}[ht]
    \centering
    \begin{tabular}{cc}
        \subfloat[log permeability realization for test case 1]{\includegraphics[width=0.45\textwidth]{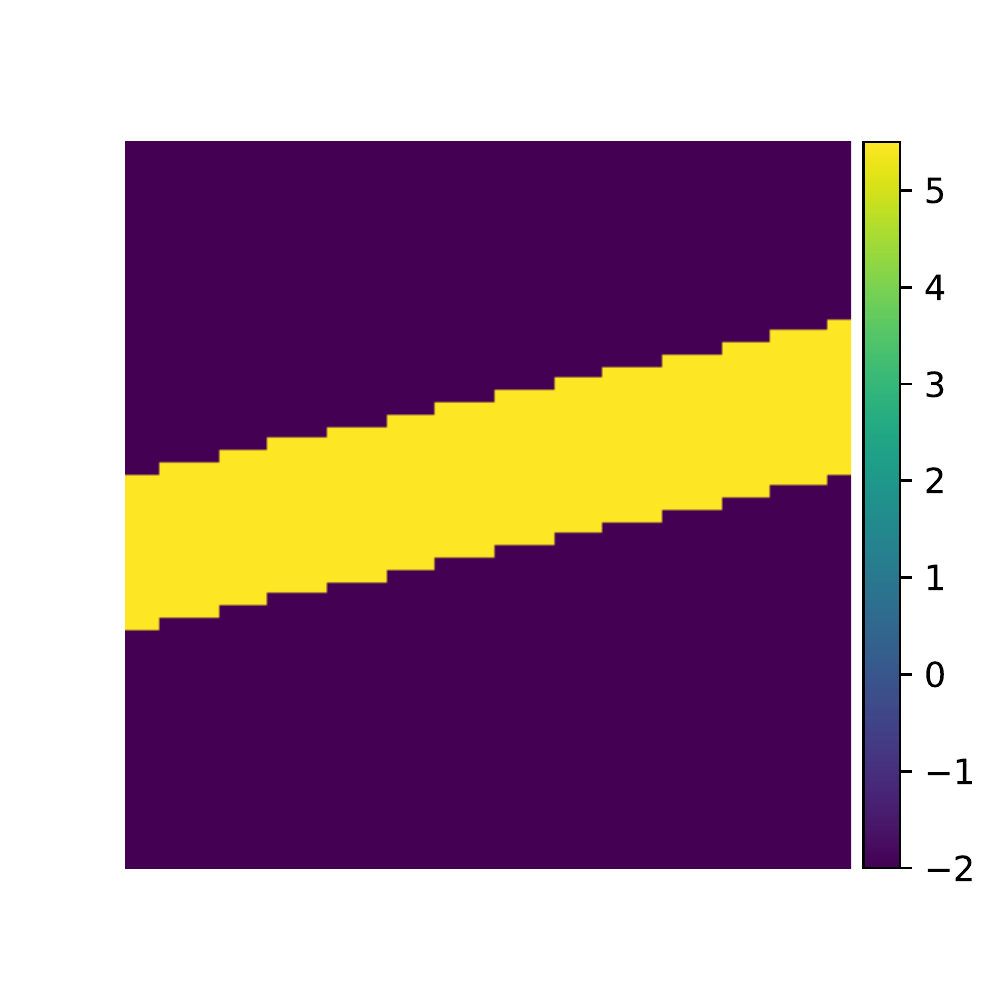}} &
        \subfloat[log permeability distribution for SPE-10 model 2 upperness case]{\includegraphics[width=0.45\textwidth]{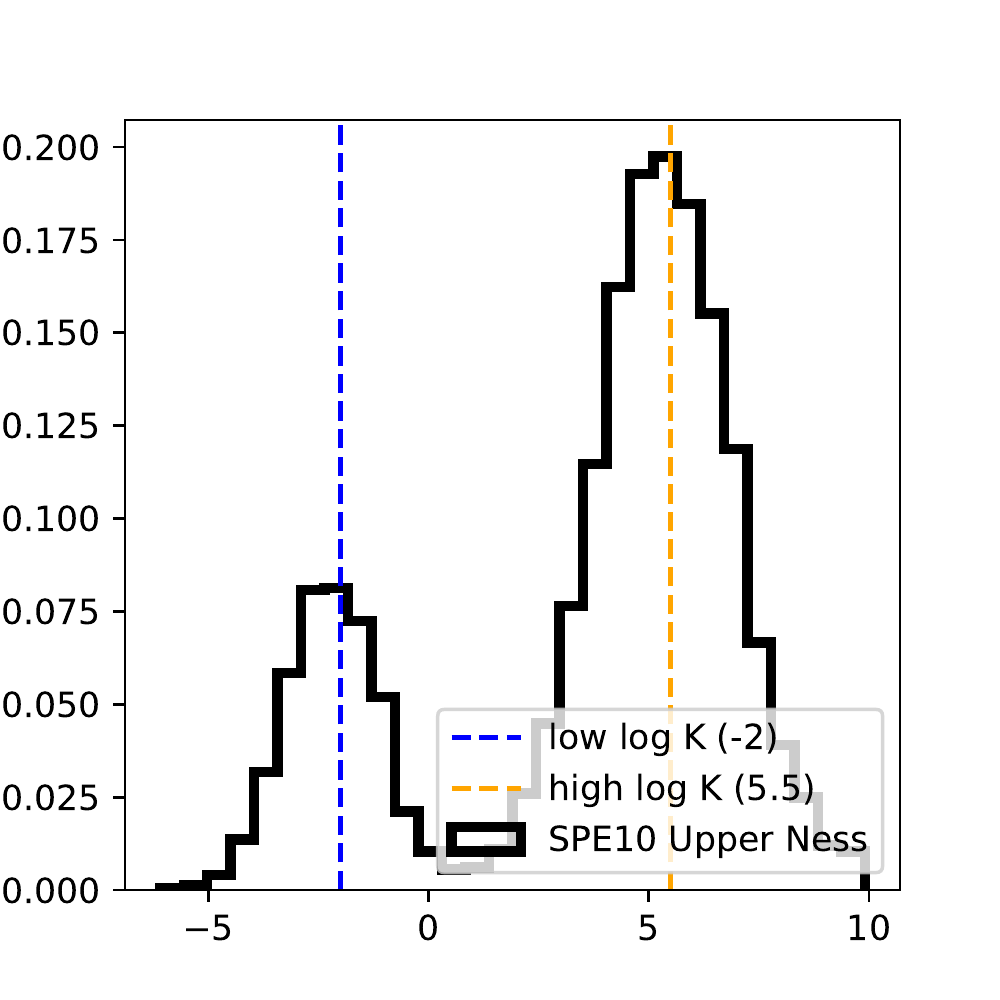}}
    \end{tabular}
    \caption{Value for low (-2) and high (5.5) log permeability in test case 1 was chosen from the SPE-10 model 2 Upperness permeability distribution peaks }
    \label{fig: case_1_k_reference}
\end{figure}

Model parameters of case studies chosen for this paper were inspired from the SPE 10 model 2 parameters. 
First test case represents channel like permeability distribution which consists of a linear high permeability channel passing through a low permeability domain. 
The values of high and low permeabilities in this distribution were selected in reference to the Upperness formation in SPE10 model 2. Figure \ref{fig: case_1_k_reference} shows the plot of Upperness log permeability distribution and an example of log permeability field in test case 1. As can be seen, the high and low log permeability values in test case 1 are chosen from the peaks of Upperness log permeability distribution.

\begin{figure}[ht]
    \centering
    \begin{tabular}{cc}
        \subfloat[log permeability realization for test case 2]{\includegraphics[width=0.45\textwidth]{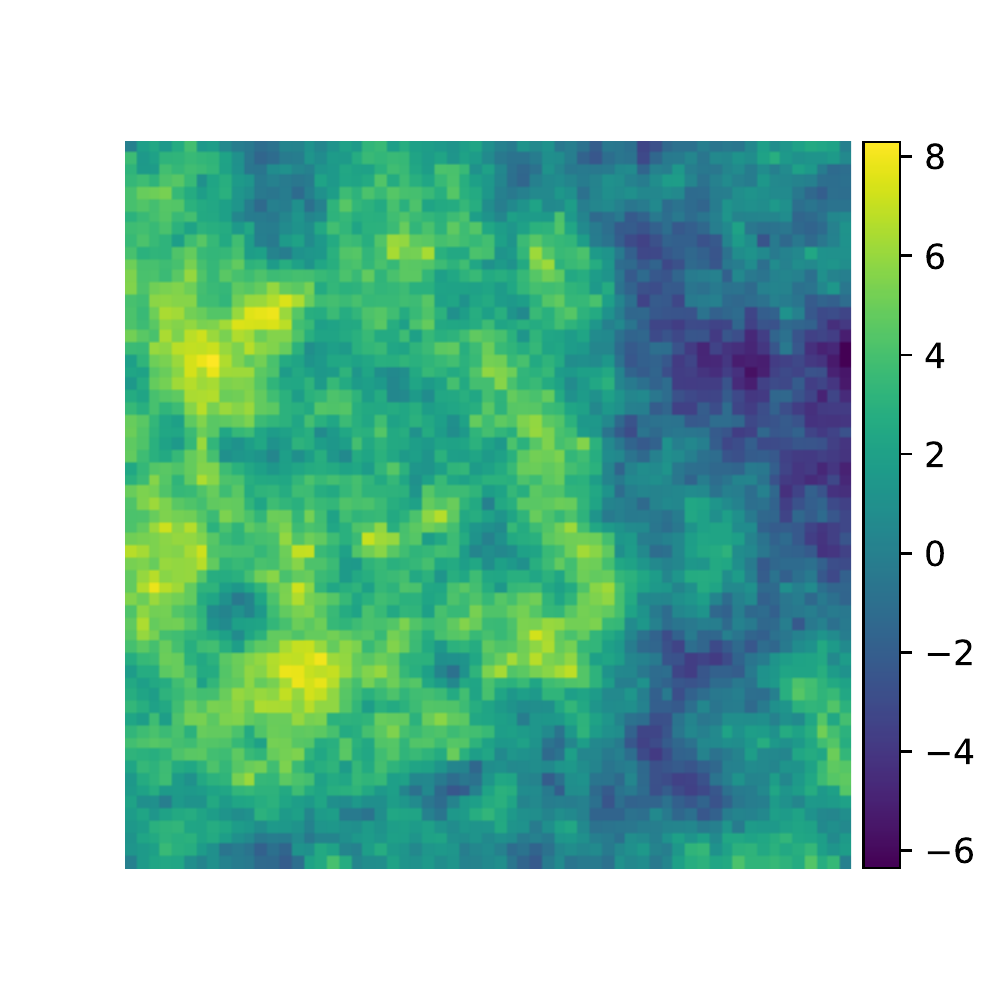}} &
        \subfloat[log permeability distribution for SPE-10 model 2 tarbert case]{\includegraphics[width=0.45\textwidth]{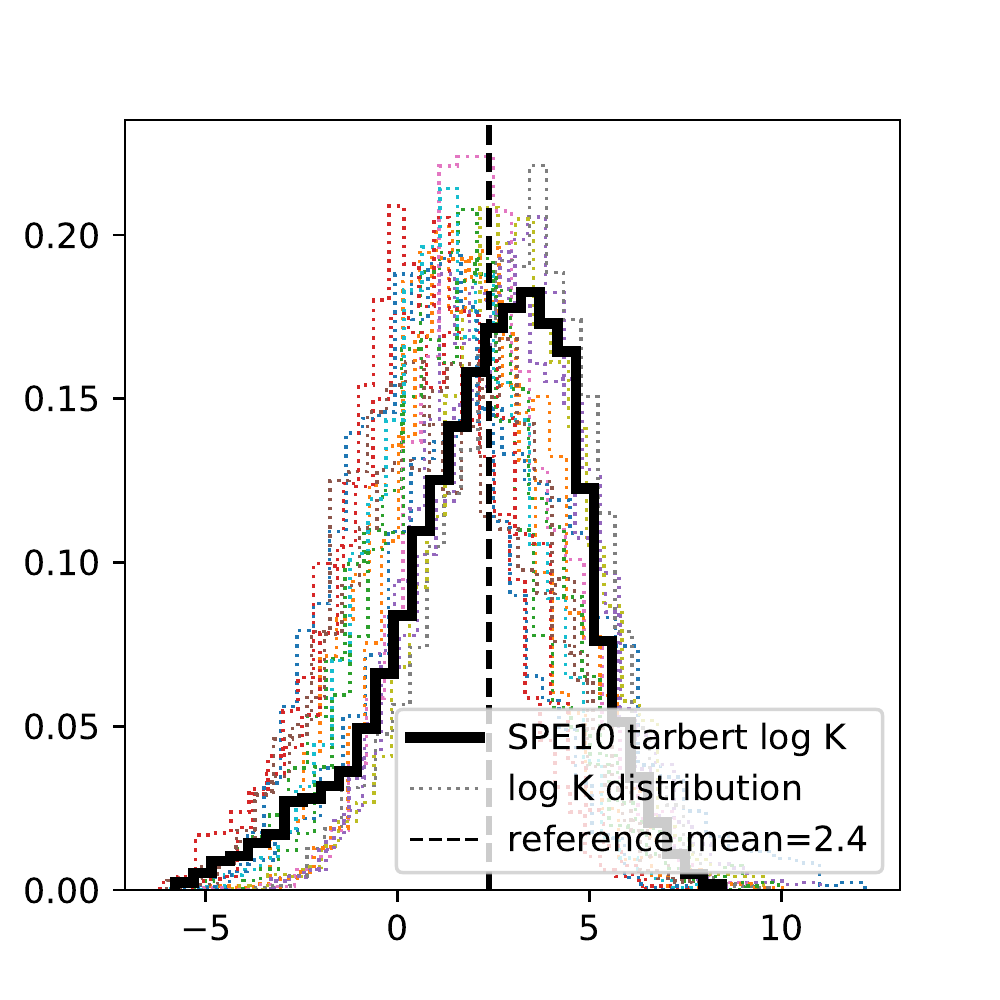}}
    \end{tabular}
    \caption{ Mean of (2.4) log permeability in test case 2 was chosen from the SPE-10 model 2 Tarbert case data. Log permeability distribution chosen in test case 2 is super-positioned with Tarbert permeability distribution (shown with thick black line) for comparison }
    \label{fig: case_2_k_reference}
\end{figure}

Second test case represents a smoother, spatially correlated permeability distribution often found in geoscience literature. The log permeability distribution is formulated with Equations \eqref{eq: g2_dist}, \eqref{eq: g2_mean}, \eqref{eq: g2_sigma} and \eqref{eq: case2_corr_len}. The correlation length $l$ was selected to be 240 ft in order to consider 20\% of the domain height. The idea is to choose a correlation length that is smaller than the quadrants of a domain. The distribution amplitude $\sigma$ and $\mu_g$ is chosen as 2.5 and 2.4, respectively. These values were chosen in order to match the log permeability distribution of test case 2 to match with Tarbert formation from SPE10 model 2. Figure \ref{fig: case_2_k_reference} shows comparison of a test case 2 log permeability field realization along with the superpositioning of test case 2 and Tarbert case permeability distribution.

\section{RL algorithm parameters} \label{app: rl_params}

\begin{figure}
    \centering
    \includegraphics[width=\columnwidth, height=0.45\textwidth]{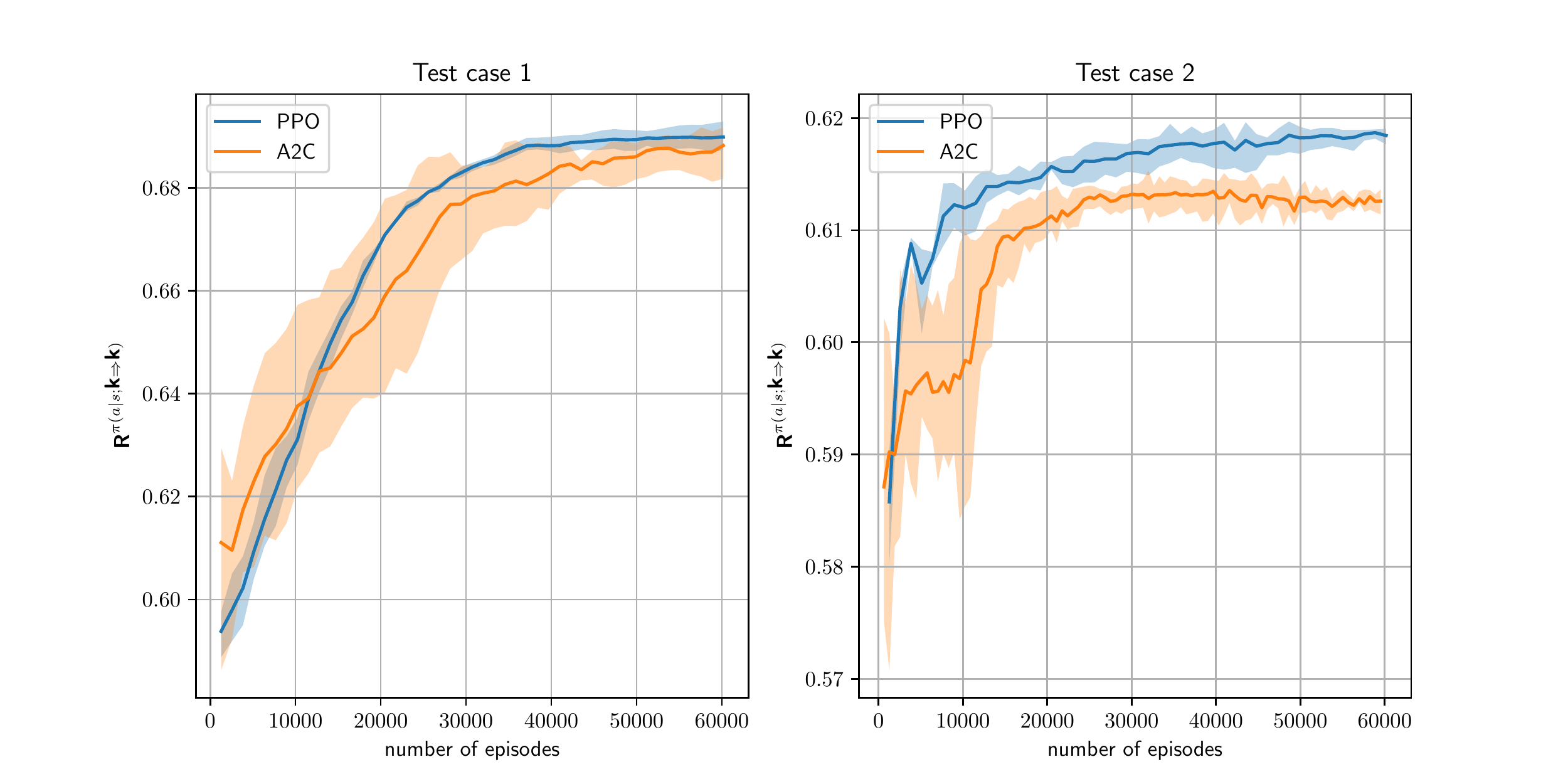}
    \caption{learning plot range over three distinct seed values for test case 1 and 2}
    \label{fig: rl_plot_range}
\end{figure}

We use stable-baselines 3 library \citep{stable-baselines3} for PPO and A2C algorithms. Parameters used for PPO and A2C are tabulated in table \ref{tab: ppo_param} and \ref{tab: a2c_param}, respectively, which were tuned using trial and error. The parameters used to obtain the frozen policy using PPO algorithms are same as those used in PPO parameters presented in these table. The DE algorithm is executed using python's SciPy library \citep{2020SciPy-NMeth}. Its parameters are delineated in table \ref{tab: de_param}.
For PPO algorithms with full state representation, same parameters (from table \ref{tab: ppo_param}) were used except the network MLP layers and learning rates: layers [3721, 4000, 2000, 800, 300, 61] and learning rate 1e-5 for test case 1 and layers [3721, 4000, 2000, 800, 300, 4] and learning rate 5e-6 for test case 2.
These parameters are tuned in order to obtain the minimum variance in the learning plots. 
Figure \ref{fig: rl_plot_range} demonstrate the spread of the learning plot for PPO and A2C for the text case 1 and 2.
The code repository for both the test cases presented in this paper can be found on the link: \url{https://github.com/atishdixit16/rl\_robust\_owc}.

\begin{table}[ht]
    \caption{PPO algorithm parameters}
    \centering
    \begin{tabular}{l l l l}
        \hline
         & Test case 1 & Test case 2\\
        \hline
        number of episodes & 60000 & 60000 \\
        number of CPUs, $N$ & 64 & 64 \\
        number of steps, $T$ & 50 & 50 \\
        mini-batch size, $M$ & 16 & 16 \\
        epochs, $K$ & 20 & 20 \\
        discount rate, $\gamma$ & 0.99 & 0.99 \\
        clip range, $\epsilon$ & 0.1 & 0.1 \\
        policy network MLP layers & [93,150,100,80,62] & [9,20,20,4] \\
        policy network activation functions  & tanh & tanh\\
        policy network optimizers  & Adam & Adam \\
        learning rate & 1e-6 & 5e-4\\
        \hline
    \end{tabular}
    \label{tab: ppo_param}
\end{table}

\begin{table}[ht]
    \caption{A2C algorithm parameters}
    \centering
    \begin{tabular}{l l l l}
        \hline
         & Test case 1 & Test case 2\\
        \hline
        number of episodes & 60000 & 60000 \\
        number of CPUs, $N$  & 64 & 32 \\
        number of steps, $T$ & 50 & 20 \\
        discount rate, $\gamma$ & 0.99 & 0.99 \\
        policy network MLP layers & [93,150,100,80,62] & [9,20,20,4] \\
        policy network activation functions & tanh & tanh \\
        policy network optimizers  & Adam & Adam \\
        learning rate & 2e-4 & 1e-4\\
        \hline
    \end{tabular}
    \label{tab: a2c_param}
\end{table}

\begin{table}[ht]
    \caption{DE algorithm parameters}
    \centering
    \begin{tabular}{l l l l}
        \hline
         & Test case 1 & Test case 2\\
        \hline
        number of CPUs  & 64 & 64 \\
        number of iterations & 750 & 750 \\
        population size & 310 & 20 \\
        recombination factor & 0.9 & 0.9 \\
        mutation factor & U(0.5,1) & U(0.5,1) \\
        \hline
    \end{tabular}
    \label{tab: de_param}
\end{table}

\clearpage
 \bibliographystyle{elsarticle-num-names} 
 \bibliography{references}





\end{document}